\documentclass[nohyperref]{article}

\usepackage{microtype}
\usepackage{graphicx}
\usepackage{subfigure}
\usepackage{booktabs} %

\usepackage{hyperref}

\usepackage[table]{xcolor} %

\usepackage[accepted]{icml2023}

\usepackage{amsmath}
\usepackage{amssymb}
\usepackage{mathtools}
\usepackage{amsthm}

\usepackage[capitalize,noabbrev]{cleveref}

\theoremstyle{plain}
\newtheorem{theorem}{Theorem}[section]
\newtheorem{proposition}[theorem]{Proposition}

\newtheorem{corollary}[theorem]{Corollary}
\theoremstyle{definition}

\theoremstyle{remark}
\newtheorem{remark}[theorem]{Remark}

\newcommand{\model}[1]{I$^2$SB}

\usepackage[textsize=tiny]{todonotes}

\usepackage{amsmath,amsfonts,bm}

\def\eqref#1{(\ref{#1})}

\def\1{\bm{1}}

\def\rd{{\textnormal{d}}}

\DeclareMathAlphabet{\mathsfit}{\encodingdefault}{\sfdefault}{m}{sl}
\SetMathAlphabet{\mathsfit}{bold}{\encodingdefault}{\sfdefault}{bx}{n}

\def\sR{{\mathbb{R}}}

\newcommand{\E}{\mathbb{E}}

\newcommand{\KL}{D_{\mathrm{KL}}}
\newcommand{\Var}{\mathrm{Var}}

\newcommand{\Cov}{\mathrm{Cov}}

\definecolor{Purple200}{HTML}{E040FB}
\definecolor{Purple400}{HTML}{D500F9}
\definecolor{DeepPurpleA400}{HTML}{651FFF}
\definecolor{Indigo400}{HTML}{3D5AFE}
\definecolor{Green400}{HTML}{00E676}
\definecolor{Green700}{HTML}{00C853}
\definecolor{Amber800}{HTML}{FF8F00}
\definecolor{Orange800}{HTML}{EF6C00}
\definecolor{DeepOrange800}{HTML}{D84315}
\definecolor{DeepOrangeA400}{HTML}{FF3D00}
\definecolor{RedA400}{HTML}{FF1744}

\newcommand*\Psihat{ \widehat{\Psi} }%

\def\gradlog{{ \nabla \log }}
\def\sbeta{{ \sqrt{\beta_t} }}
\def\sigmabar{{ \bar{\sigma} }}

\def\barW{{ \overline{W} }}

\newcommand{\norm}[1]{\lVert#1\rVert}

\def\dt{{ \mathrm{d} t }}
\def\ds{{ \mathrm{d} s }}

\def\calA{{\cal A}}
\def\calB{{\cal B}}

\def\calN{{\cal N}}
\def\calO{{\cal O}}

\def\calU{{\cal U}}

\newcommand{\fracpartial}[2]{\frac{\partial #1}{\partial  #2}}

\newcommand{\br}[1]{\left[#1\right]}
\newcommand{\pr}[1]{\left(#1\right)}
\newcommand{\T}{\top}

\usepackage{lipsum}

\usepackage{color,soul}
\definecolor{azure}{rgb}{0.0, 0.5, 1.0}

\newcommand{\markgreen}[1]{{\color{green!50!black} #1}}
\newcommand{\markred}[1]{{\color{red!80!black} #1}}
\newcommand{\markblue}[1]{{\color{azure!80!black} #1}}

\colorlet{llgray}{lightgray!40}

\newcommand\numberthis{\addtocounter{equation}{1}\tag{\theequation}}
\newcommand{\eg}{{\ignorespaces\emph{e.g.,}}{ }}
\newcommand{\ie}{{\ignorespaces\emph{i.e.,}}{ }}

\usepackage{aligned-overset}
\allowdisplaybreaks
\usepackage{cancel}

\usepackage{enumitem}
\usepackage{multirow}

\newcommand{\specialcell}[2][c]{%
  \begin{tabular}[#1]{@{}c@{}}#2\end{tabular}}

\usepackage{wrapfig}

\hypersetup{
    colorlinks,
    linkcolor={black},
    citecolor={blue!50!black},
    urlcolor={blue!50!black}
}

\icmltitlerunning{Image-to-Image Schr\"odinger Bridge}

\begin{document}

\twocolumn[
\icmltitle{\model{}: Image-to-Image Schr\"odinger Bridge}

\icmlsetsymbol{equal}{$\dagger$}

\begin{icmlauthorlist}
\icmlauthor{Guan-Horng Liu~\textsuperscript{*}}{gatech}
\icmlauthor{Arash Vahdat}{nvidia}
\icmlauthor{De-An Huang}{nvidia}
\icmlauthor{Evangelos A. Theodorou}{gatech} \\
\icmlauthor{Weili Nie}{equal,nvidia}
\icmlauthor{Anima Anandkumar}{equal,nvidia,caltech}
\end{icmlauthorlist}

\icmlaffiliation{gatech}{Georgia Institute of Technology}
\icmlaffiliation{nvidia}{NVIDIA}
\icmlaffiliation{caltech}{California Institute of Technology}

\icmlcorrespondingauthor{}{ghliu@gatech.edu}

\icmlkeywords{Machine Learning, ICML}

\vskip 0.3in
]

\printAffiliationsAndNotice{\textsuperscript{*}Work done in part as a research intern at NVIDIA. \icmlEqualAdvising.} %

\begin{abstract}

We propose {Image-to-Image Schr\"odinger Bridge (\model{})}, a new class of conditional diffusion models
that directly learn the nonlinear diffusion processes between two given distributions.
These \emph{diffusion bridges} are particularly useful for image restoration, as the degraded images
are structurally informative priors for reconstructing the clean images.
\model{} belongs to a tractable class of Schr\"odinger bridge, the nonlinear extension to score-based models, whose marginal distributions can be computed analytically given boundary pairs.
This results in a simulation-free framework for nonlinear diffusions, where the \model{} training becomes scalable by adopting practical techniques used in standard diffusion models.
We validate \model{} in solving various image restoration tasks, including inpainting, super-resolution, deblurring, and JPEG restoration on ImageNet 256$\times$256
and show that \model{} surpasses standard conditional diffusion models with more interpretable generative processes.
Moreover, \model{} matches the performance of inverse methods that additionally require the knowledge of the corruption operators.
Our work opens up new algorithmic opportunities for developing efficient nonlinear diffusion models on a large scale.
{Project page and codes}: \url{https://i2sb.github.io/}.

\end{abstract}

\def\qddpm{{ q^\text{\eqref{eq:fsgm}} }}
\def\qsb{{ q^\text{\eqref{eq:sb-sde}} }}
\def\cin{{ c^\text{in}_t }}
\def\cout{{ c^\text{out}_t }}
\def\cskip{{ c^\text{skip}_t }}

\section{Introduction}

\begin{figure}[t]
    \begin{center}
    \centerline{\includegraphics[width=\columnwidth]{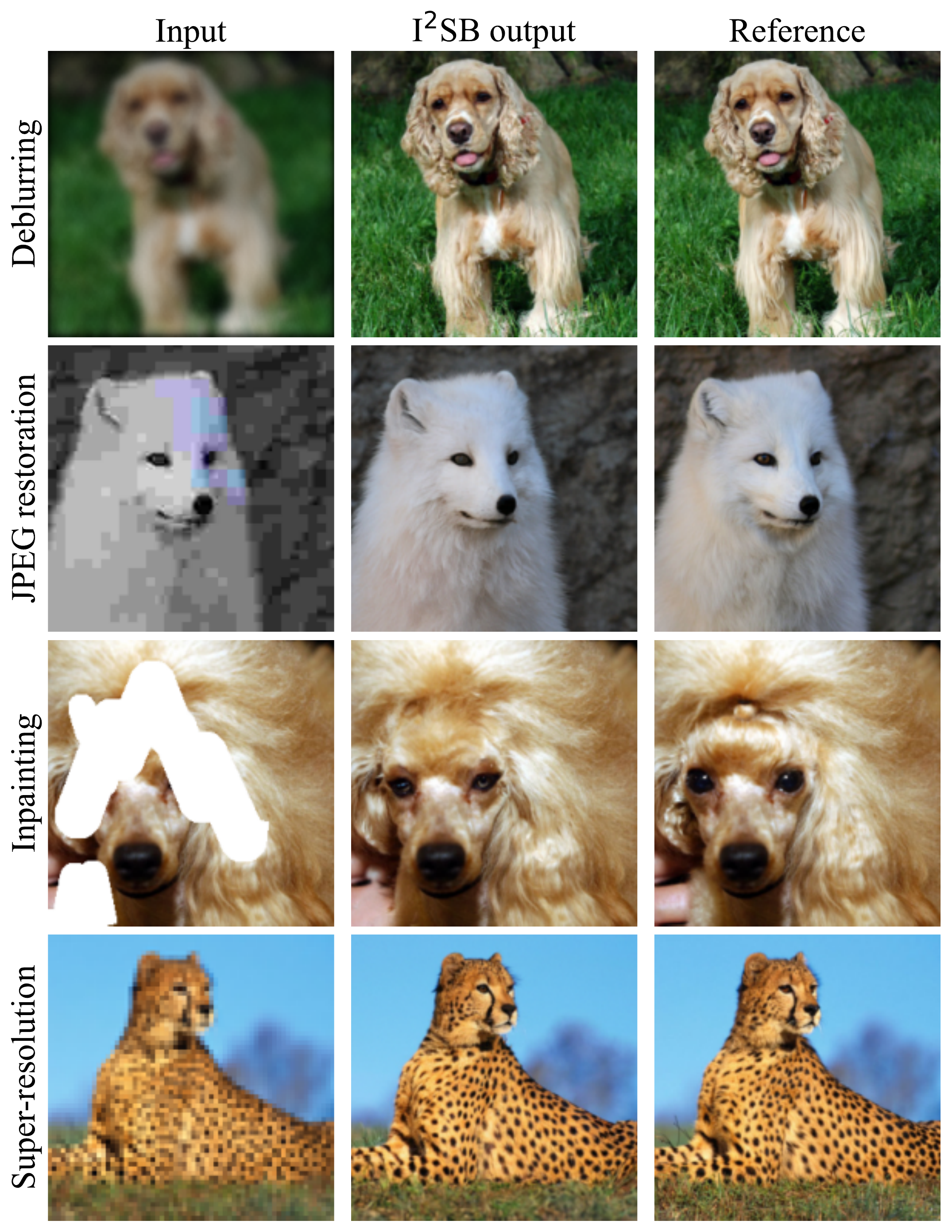}}
    \vskip -0.15in
    \caption{
        Outputs of our proposed \textbf{Image-to-Image Schr\"odinger Bridge (\model{})} on ImageNet 256$\times$256 validation set for various image restoration tasks.
    }
    \label{fig:cdb}
    \end{center}
    \vskip -0.3in
\end{figure}

Image restoration is a crucial problem in vision and image processing with applications in optimal filtering \citep{motwani2004survey}, data compression \citep{wallace1991jpeg}, adversarial defense \citep{nie2022diffusion}, and safety-critical systems such as medicine and robotics \citep{song2021solving,li2021deep}.
Common image restoration tasks
are known to be ill-posed \citep{banham1997digital,richardson1972bayesian} and
typically solved by modern data-driven approaches with conditional generation \citep{mirza2014conditional,khan2022transformers}, \ie by learning to sample the underlying (clean) data distribution given the degraded distribution.

Diffusion and score-based generative models (SGMs; \citet{sohl2015deep,song2020score}) have emerged as powerful conditional generative models with their remarkable successes in synthesizing high-fidelity data \citep{dhariwal2021diffusion,rombach2022high,vahdat2021score}.
These models rely on progressively diffusing data to noise, and learning the score functions (often parameterized by neural networks) to reverse the processes \citep{anderson1982reverse};
the reversed processes enable generation from noise to data.
\citet{saharia2021image,saharia2022palette} show that these generative processes can be adopted for image restoration by feeding degraded images as extra inputs to the score network so that the processes are biased toward the corresponding intact images.
Alternatively, when the mapping between clean and degraded images is known,
the tasks can be reformulated as inverse problems that restore the underlying clean signal from the degraded measurement, based on the diffusion priors~\citep{kawar2022denoising,kawar2022jpeg,wang2022zero}.

Notably, all of the aforementioned diffusion models for image restoration begin their generative denoising processes with Gaussian white noise, which has little or no structural information of the clean data distribution.
Despite arising naturally from unconditional generation, it remains unclear whether this default setup best suits image-to-image translation problems especially like image restoration, where the degraded images are much more structurally informative compared to random noise.

An alternative that better leverages the problem structure is to directly start the generative processes from degraded images, and build diffusion \emph{bridges} between clean and degraded data distributions.
This shares similarity with image-to-image translation GANs~\citep{zhu2017unpaired,huang2018multimodal}.
Constructing these diffusion \emph{bridges} often necessitates a new computational framework for reversing general diffusion processes.
It has been recently explored in Schr\"odinger bridge (SB; \citet{de2021diffusion,chen2021likelihood}), a generalized nonlinear score-based model which defines optimal transport between two arbitrary distributions and generalizes beyond Gaussian priors.

\begin{figure}[t]
    \begin{center}
    \centerline{\includegraphics[width=0.85\columnwidth]{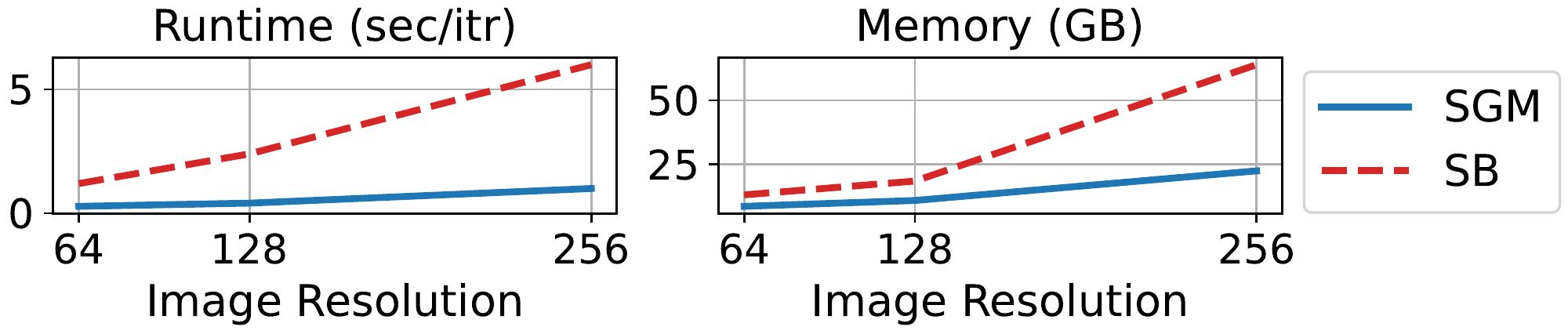}}
    \vskip -0.13in
    \caption{
        Complexity of SGM and SB \citep{chen2021likelihood} On 256$\times$256 resolution,
        SB is 6$\times$ slower and consumes 3$\times$ memory.
    }
    \label{fig:complexity}
    \end{center}
    \vskip -0.25in
\end{figure}

\begin{figure}[t]
    \begin{center}
    \centerline{\includegraphics[width=0.9\columnwidth]{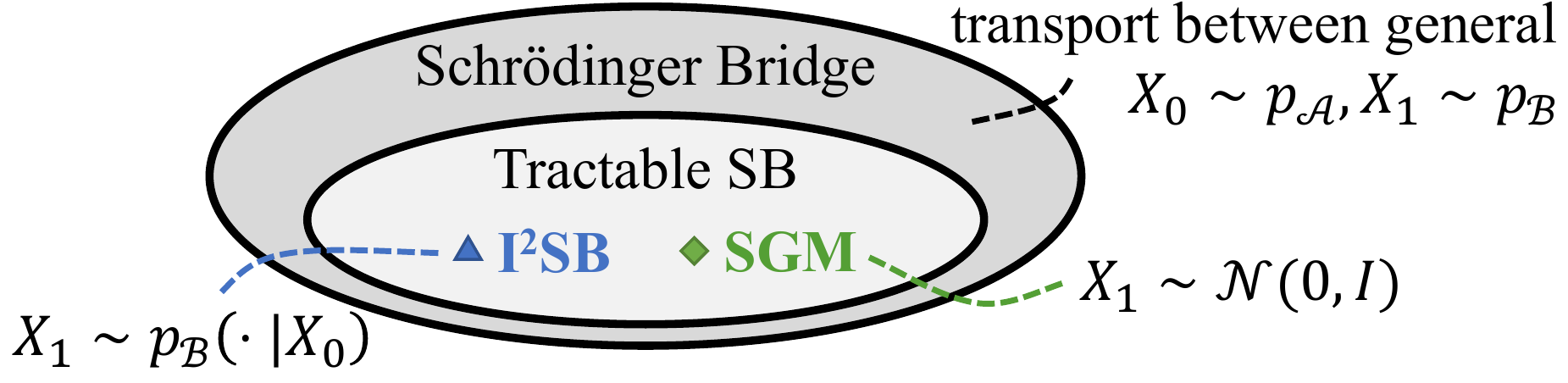}}
    \vskip -0.1in
    \caption{
        \textbf{\model{}} belongs to a tractable class of SB that shares the same computational framework of SGM and rebases the terminal distribution beyond simple Gaussian priors.
    }
    \label{fig:venn}
    \end{center}
    \vskip -0.2in
\end{figure}

Despite the mathematical generalization,
computational frameworks for solving SB \citep{chen2021stochastic}
have been developed independently (hence distinctly) from how diffusion models are typically trained.
This makes SB computationally unfavorable compared to its score-based counterpart especially in high-dimensional regimes (see \cref{fig:complexity}), where SB is known to suffer from, \eg discretization error \citep{de2021diffusion}, high variance \citep{chen2021likelihood}, or even divergence \citep{fernandes2021shooting}.
It remains an open question whether SB can be made practical for learning complex nonlinear diffusions on a large scale.

In this work, we propose \textbf{Image-to-Image Schr\"odinger Bridge (\model{})}, a sub-class of SB with nonlinear diffusion models that share the same computational framework used in standard score-based models.
Consequently, practical techniques from diffusion models for learning high-dimensional data distributions \citep{karras2022elucidating,song2020improved} can be adopted to train nonlinear diffusions.
This is achieved by exploiting the linear structure hidden in the nonlinear coupling of SB
to construct tractable SBs for transporting between individual clean images and their corresponding degraded distributions, \ie \model{}.
We show that the marginal distributions of \model{} admit analytic solutions given boundary pairs (\ie clean and degraded image pairs), thereby yielding a simulation-free framework that avoids unfavorable complexity \citep{chen2021likelihood}.
Furthermore, we demonstrate that \model{} can be simulated at test time using DDPM \citep{ho2020denoising}.
Finally, we characterize in how \model{} reduces to an optimal transport ODE \citep{peyre2019computational} when the diffusion vanishes, strengthening the algorithmic connection among dynamic generative models.

\begin{figure}[t]
    \vskip 0.02in
    \begin{center}
    \centerline{\includegraphics[width=\columnwidth]{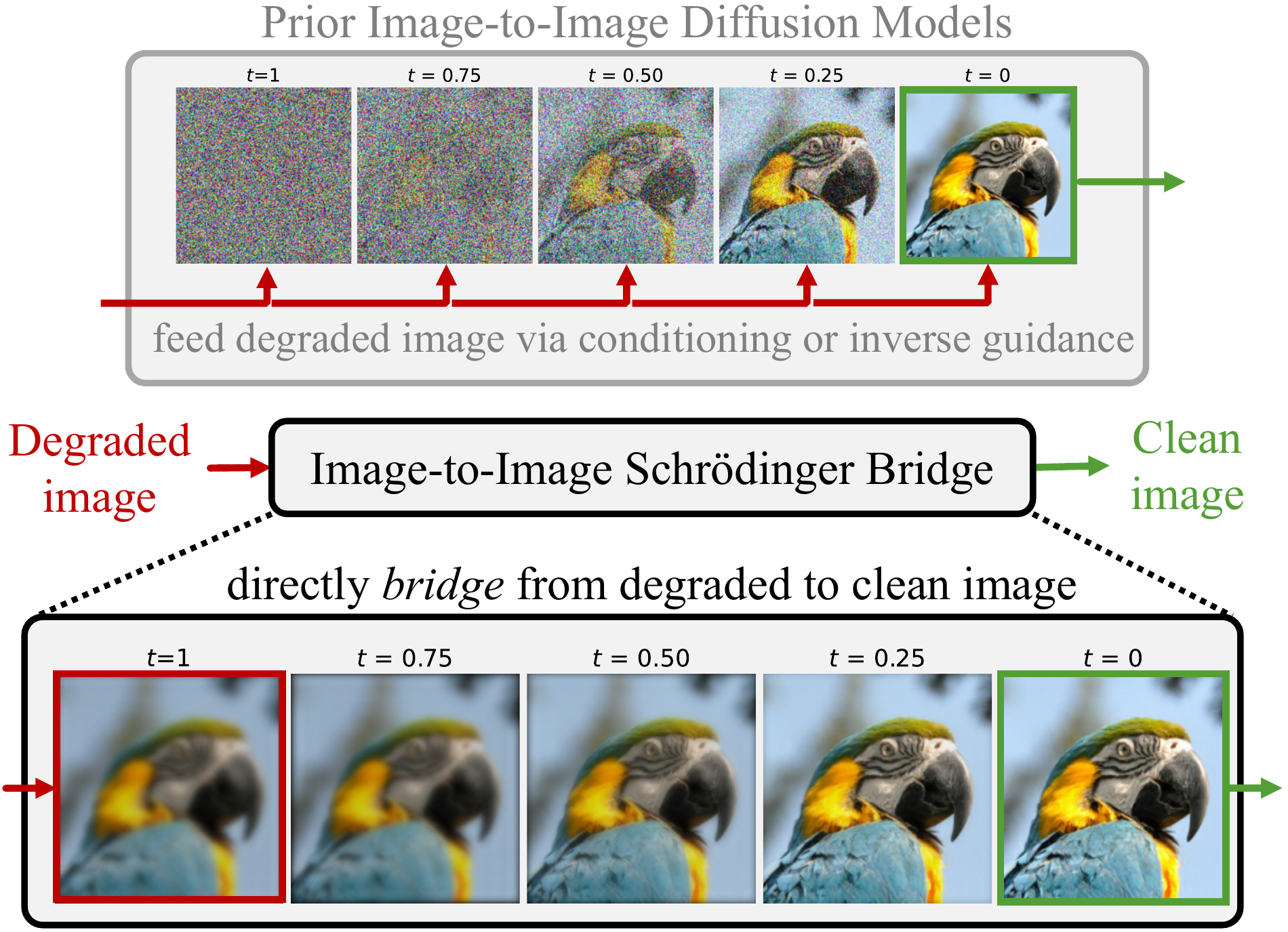}}
    \vskip -0.07in
    \caption{
        Illustration of \textbf{I$^2$SB}. Rather than generating images from random noise as in {prior diffusion models},
        \model{} directly learns the diffusion bridges between degraded and clean distributions, yielding more interpretable generation effective for image restoration.
    }
    \label{fig:diagram}
    \end{center}
    \vskip -0.2in
\end{figure}

We validate our method in many image restoration tasks including super-resolution, deblurring, inpainting, and JPEG restoration on ImageNet 256$\times$256 \citep{deng2009imagenet}; see \cref{fig:cdb}.
Through extensive experiments, we show that \model{} surpasses standard conditional diffusion models \citep{saharia2022palette} and matches diffusion-based inverse models \citep{kawar2022denoising,kawar2022jpeg} \emph{without} exploiting the corruption operators.
With these more interpretable generative processes, \model{} enjoys little or no performance drops as the number of function evaluation (NFE) decreases.

In summary, we present the following contributions.
\vspace{-8pt}
\begin{itemize}[leftmargin=10pt]
    \item We introduce \model{}, a new class of conditional diffusion models that learns fully nonlinear diffusion bridges between two domain distributions.
    \item We build \model{} on a simulation-free computational framework that adopts scalable techniques from standard diffusion models to train nonlinear diffusion processes.
    \item \model{} sets new records in many restoration tasks, including super-resolution, deblurring, inpainting, and JPEG restoration. It yields more interpretable generation and enjoys little performance drops as the NFE decreases.
\end{itemize}

\section{Preliminaries}

\paragraph*{Notation:}
Let $X_t \in \sR^d$ be a $d$-dimensional stochastic process indexed by $t\in[0,1]$
and denote the discrete time step by $0=t_0 < \cdots t_n \cdots < t_N = 1$, we shorthand $X_n \equiv X_{t_n}$.
The Wiener process and its reversed counterpart \citep{anderson1982reverse} are denoted by $W_t, \barW_t \in \sR^d$.
The identity matrix is denoted by $I \in \mathbb{R}^{d \times d}$.

\subsection{Score-based Generative Model (SGM)}

SGM \citep{sohl2015deep,ho2020denoising,song2020score} is an emerging class of dynamic generative models that,
given data $X_0$ sampled from some domain $p_\calA$, %
constructs stochastic differential equations (SDEs),
\begin{align} \label{eq:fsgm}
    \rd X_t = f_t(X_t) \dt + \sbeta \rd W_t, %
\end{align}
whose terminal distributions at $t=1$ approach an approximate Gaussian, \ie $X_1 \sim \calN(0,I)$.
This is achieved by properly choosing the diffusion $\beta_t\in \mathbb{R}$ and setting the base drift $f_t$ linear in $X_t$.
It is known that reversing \eqref{eq:fsgm} yields another SDE traversing backward in time \citep{anderson1982reverse}:
\begin{align} \label{eq:rsgm}
    \rd X_t = [f_t - \beta_t \gradlog~p(X_t,t)] \dt + \sbeta \rd \barW_t,
\end{align}
where $p(\cdot,t)$ is the marginal density of \eqref{eq:fsgm} at time $t$
and $\gradlog~p$ is its \emph{score function}.
The SDE \eqref{eq:rsgm} is known as the ``reversed process of \eqref{eq:fsgm}'' in the sense that
its path-wise measure equals almost surely to the one induced by \eqref{eq:fsgm};
thus, the two SDEs also share the same marginal densities.

In practice, given a tuple $(X_0,t,X_t)$ where $X_0\sim p_\calA$, $t\sim\calU([0,1])$, and $X_t$ sampled analytically from \eqref{eq:fsgm}, one can parameterize $\epsilon(X_t,t; \theta)$ with, \eg U-Net \citep{ronneberger2015u}, and regress its output w.r.t. the rescaled version of denoising score-matching objective \citep{vincent2011connection},
\begin{align} \label{eq:dsm}
    \norm{ \epsilon(X_t,t; \theta) - \sigma_t \gradlog~p(X_t,t | X_0)},
\end{align}
where $\gradlog~p(X_t,t | X_0)$ can be computed analytically %
and $\sigma_t^2$ is the variance of $X_t|X_0$, induced by \eqref{eq:fsgm}, that rescales the regression target to unit variance \citep{ho2020denoising}.

Other advanced parameterizations that better account for practical training \citep{karras2022elucidating} have also been explored recently.
Importantly, they all produce some ways to predict intact data at $t=0$ from the network outputs. In other words, the mapping $\epsilon(X_t,t; \theta) \mapsto X_0^\epsilon$ is readily available once $\epsilon$ is trained.\footnote{
    In all cases, we can write $X_t = a_t X_0 + b_t \epsilon$ for some $a_t,b_t\in \sR$ depending on \eqref{eq:fsgm} so that the mapping can be defined as $X_0^\epsilon := (X_t - b_t \epsilon_\theta)/a_t$ given any trained $\epsilon_\theta$.
}
With this, popular samplers like DDPM \citep{ho2020denoising} can be written compactly as recursive posterior sampling:
\begin{align}
    \label{eq:ddpm}
    X_n \sim p(X_n | {X}_0^\epsilon, X_{n+1}), \quad X_N \sim \calN(0,I).
\end{align}

\subsection{Schr\"odinger Bridge (SB)} \label{sec:sb-theory}

SB \citep{schrodinger1932theorie,leonard2013survey} is an {entropy-regularized optimal transport} model that considers the following forward and backward SDEs:
\begin{subequations}
    \begin{align}
        \rd X_t &= [f_t + \beta_t~\gradlog {\Psi}(X_t, t) ] \dt + \sbeta\rd W_t, \label{eq:fsb}
        \\
        \rd X_t &= [f_t - \beta_t~\gradlog \Psihat(X_t, t) ] \dt + \sbeta\rd \barW_t, \label{eq:rsb}
    \end{align} \label{eq:sb-sde}%
\end{subequations}
where $X_0 \sim p_\calA$ and $X_1 \sim p_\calB$ are drawn from boundary distributions in two distinct domains. The functions $\Psi, \Psihat \in C^{2,1}(\sR^d,[0,1])$ are time-varying energy potentials that
solve the following coupled PDEs,
\begin{subequations}\label{eq:sb-pde3}
\begin{align}
   & \begin{cases}
  \fracpartial{\Psi(x,t)}{t}    = - \nabla \Psi^\T f - \frac{1}{2} \beta \Delta \Psi \\[3pt]
  \fracpartial{\Psihat(x,t)}{t} = - \nabla \cdot (\Psihat f) + \frac{1}{2} \beta \Delta \Psihat
  \end{cases} \label{eq:sb-pde}
  \\
   \text{s.t. }
  \Psi(x,0) &  \Psihat(x,0){=}p_\calA(x),
      \Psi(x,1) \Psihat(x,1){=}p_\calB(x) \label{eq:sb-pde2}
\end{align}
\end{subequations}
In this case, the path measure induced by SDE \eqref{eq:fsb} equals almost surely to the one induced by SDE \eqref{eq:rsb}, similar to SDEs (\ref{eq:fsgm},\ref{eq:rsgm}).
Hence, their marginal densities, denoted by $q(\cdot,t)$ hereafter, are also equivalent.

\textbf{SGM as a Special Case of SB$\quad$}
It is known that SB generalizes SGM to nonlinear structure \citep{chen2021likelihood}.
Indeed, the SDEs between SGM (\ref{eq:fsgm},\ref{eq:rsgm}) and SB \eqref{eq:sb-sde} differ only by the additional nonlinear forward drift $\gradlog\Psi$, which allows the processes to transport samples beyond Gaussian priors.
In such cases, the backward drift $\gradlog\Psihat$ is no longer the score function of \eqref{eq:fsb}, yet they relate to each other via the Nelson's duality \citep{nelson1967dynamical}
\begin{equation}
    \label{eq:corrector}
    \begin{split}
        {\Psi}(x,t) {\Psihat}(x,t) &= q(x,t) \\
        {\Rightarrow}
        \gradlog {\Psi}(x,t) -\gradlog~q(x,t) &= -\gradlog {\Psihat}(x,t).
    \end{split}
\end{equation}
One can verify that reversing \eqref{eq:fsb} yields
\begin{align} \label{eq:rr}
    \rd X_t
    {=} [f_t + \beta_t~\gradlog \Psi - \beta_t \gradlog~q] \dt {+} \sbeta\rd \barW_t,
\end{align}
which indeed equals \eqref{eq:rsb} after substituting \eqref{eq:corrector}.
Hence, \eqref{eq:rsb} reverses the nonlinear forward SDE \eqref{eq:fsb}, and vice versa.

\section{Image-to-Image Schr\"odinger Bridge (\model{})} \label{sec:cdb}

We propose a tractable class of SB that directly constructs diffusion bridges between two domains, making it suitable for image-to-image translation such as image restoration.
All proofs are left to \cref{sec:proof} due to space constraint.

\subsection{Mathematical Framework} \label{sec:sb-sgm}

\textbf{Solving SB using SGM Framework$\quad$}
Despite the fact that SB generalizes SGM in theory,
numerical methods for SB and SGM have been developed independently on distinct computational frameworks.
Due to the coupling constraints in \eqref{eq:sb-pde2},
modern SB models often adopt iterative projection methods \citep{kullback1968probability,chen2021stochastic}, which have unfavorable complexity as the dimension grows (see \cref{fig:complexity}).
It is unclear whether practical techniques in the SGM computational framework can be transferred to efficiently learn nonlinear diffusions.

Let us reexamine the SB theory in detail, but this time through the computational framework of SGM. Notice that
\vspace{-8pt}
\begin{itemize}[leftmargin=10pt]
    \item The nonlinear drifts in \eqref{eq:sb-sde} resemble the score function in \eqref{eq:rsgm} when we view $\Psi(\cdot,t)$ and $\Psihat(\cdot,t)$ as the densities.
    \item \cref{eq:sb-pde} gives the solution to the Fokker-Plank equation \citep{risken1996fokker} that characterizes the marginal density induced by the \textit{linear} SDE in \eqref{eq:fsgm}.
\end{itemize}
\vspace{-4pt}
With these, we can reformulate PDEs \eqref{eq:sb-pde3} in a manner that makes SB more compatible with the SGM framework:
\begin{theorem}[Reformulating SB drifts as score functions]
    \label{thm:1}
    When the Schr\"odinger systems \eqref{eq:sb-pde3} hold,
    $\gradlog \Psihat(X_t,t)$ and $\gradlog \Psi(X_t,t)$ are the score functions
    of the following linear SDEs, respectively:
    \begin{subequations}\label{eq:fp-sde}
      \begin{align}
          \rd X_t &= f_t(X_t) \dt + \sqrt{\beta_t}~\rd W_t, \quad X_0 \sim \Psihat(\cdot, 0), \label{eq:fp-fsde}
          \\
          \rd X_t &= f_t(X_t) \dt + \sqrt{\beta_t}~\rd \barW_t, \quad X_1 \sim \Psi(\cdot, 1). \label{eq:fp-rsde}
      \end{align}
      \end{subequations}
    \vspace{-12pt}
\end{theorem}
\cref{thm:1} suggests that the backward drift $\gradlog \Psihat$ in SDE~\eqref{eq:rsb} that transports samples from $p_\calB$ to $p_\calA$ can also be used to reverse the forward SDE \eqref{eq:fp-fsde}.  Crucially, the above linear SDEs \eqref{eq:fp-sde} have different boundary distributions from nonlinear SDEs \eqref{eq:sb-sde}.
Essentially, the nonlinearity of $\gradlog \Psihat$---as the combination of the nonlinear forward drift and its score function (\textit{c.f.} \eqref{eq:corrector})---is absorbed into the initial condition $\Psihat(\cdot, 0)$, leaving it compactly as the score function of another linear SDE.
Hence, if we can draw samples from $X_0 \sim  \Psihat(\cdot, 0)$, we can parameterize $\gradlog \Psihat$ with the score network and apply practical techniques from SGM to learn $\gradlog \Psihat$.
Similar reasoning applies to $\gradlog \Psi$. %

\begin{figure}[t]
    \vskip 0.05in
    \begin{center}
    \centerline{\includegraphics[width=\columnwidth]{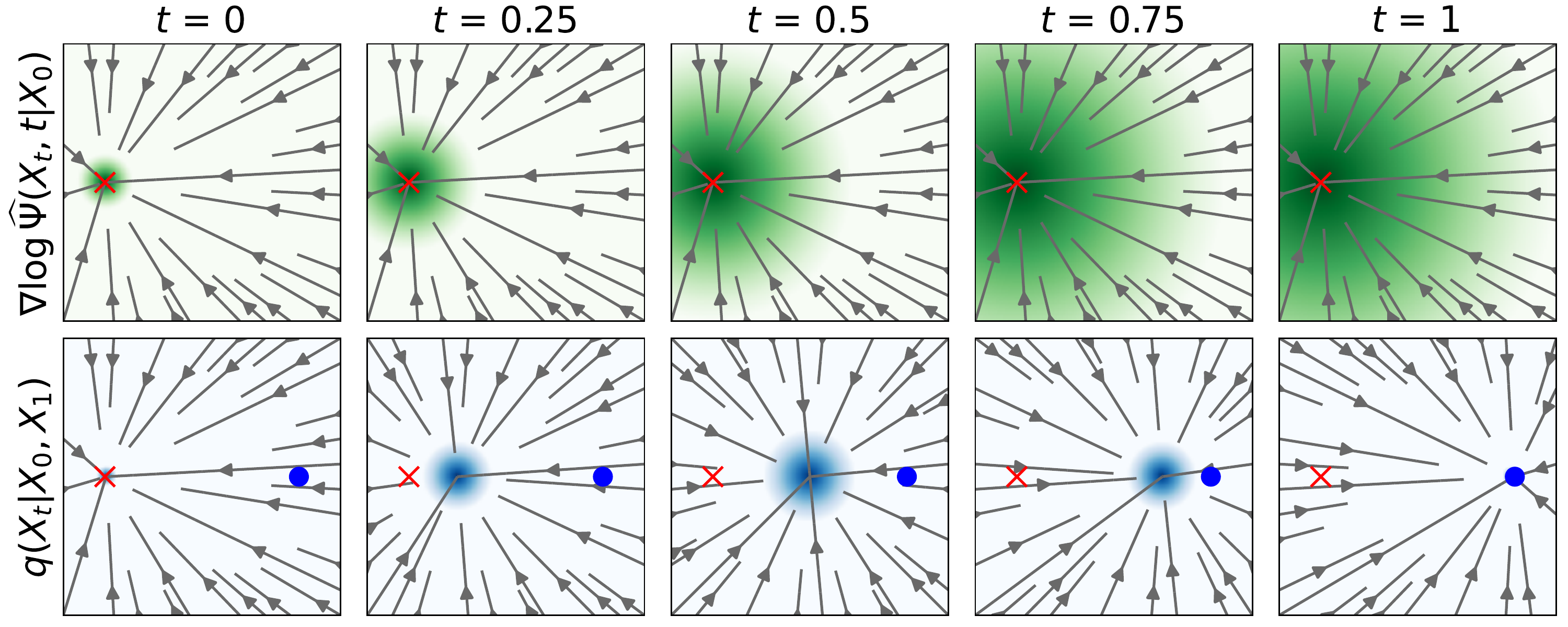}}
    \vskip -0.1in
    \caption{
        \textbf{Top:}
            The backward drift $\gradlog \Psihat(X_t,t|X_0{=}a)$ transporting $p_\calB$ toward $a$ (denoted {\color{red}$\times$}) corresponds to the score function of a \markgreen{tractable Gaussian (density marked {green})}.
        \textbf{Bottom:}
            Simulating the backward SDE \eqref{eq:rsb} with this $\gradlog \Psihat$ yields a \markblue{diffusion bridge (density marked blue)} between $X_0=a$ and $X_1\sim p_\calB$, whose mean corresponds to the optimal transport path. %
    }
    \label{fig:coro2}
    \end{center}
    \vskip -.25in
\end{figure}

\textbf{A Tractable Class of SB$\quad$}
\cref{thm:1} is encouraging yet not immediately useful
as the boundaries $\Psihat(\cdot, 0)$ and $\Psi(\cdot, 1)$ remain intractable
due to the couplings in \eqref{eq:sb-pde2}.
Below, we present a tractable case that eliminates one of the couplings.
\begin{corollary}[Tractable SB with the Dirac delta boundary] \label{coro:2}
    Let $p_\calA(\cdot) := \delta_{a}(\cdot)$
    be the Dirac delta distribution centered at $a \in \sR^d$.
    Then,
    the initial distributions in (\ref{eq:fp-sde})
    are given by
    \begin{align}
      \Psihat(\cdot, 0) = \delta_{a}(\cdot), \quad
      \Psi(\cdot, 1) = \tfrac{p_\calB}{\Psihat(\cdot, 1)}. \label{eq:coro2}
    \end{align}
    \vspace{-16pt}
\end{corollary}
Comparing \eqref{eq:coro2} to \eqref{eq:sb-pde2}, it is clear that
\cref{coro:2} breaks the dependency on $\Psi$ for solving $\Psihat(x,0)$.
Intuitively, the optimal\footnote{
    The optimality is w.r.t. minimum energy; see \cref{sec:sb-intro}.
} backward drift driving the reverse process of \eqref{eq:fp-fsde} to the Dirac delta $\delta_{a}(\cdot)$
always flows toward $a$, regardless of $p_\calB$; see \cref{fig:coro2}.
The Dirac delta assumption also implicitly appears in the denoising objective \eqref{eq:dsm},
which first computes the target $\gradlog~p(X_t,t|X_0{=}a)$ for each data point $a$, as the score between $\delta_{a}(\cdot)$ and Gaussian, then averages over $X_0 {\sim} p_\calA$.
In this vein,
\cref{coro:2} adopts the same boundary $\delta_{a}(\cdot)$ on one side and generalizes the other side from Gaussian to arbitrary $p_\calB$.
Indeed, we show in \cref{sec:proof} that when $p_\calB = \Psihat(\cdot, 1) \approx \calN(0,I)$, the forward drift vanishes with $\Psi(\cdot,t) = 1$, reducing the framework to SGM.

Although the singularity of $\delta_{a}(\cdot)$ may hinder generalization beyond training samples, in practice, the score network generalizes well to unseen samples from the same distributions, for both SGM and our \model{}, partly due to the strong generalization ability of neural networks~\citep{zhang2021understanding}.

To summarize, our theories suggest an efficient pipeline for training $\gradlog \Psihat$ without dealing with the intractability of reversing the nonlinear forward drift. By formulating a tractable SB compatible with the SGM framework,
we get both mathematical soundness and computational efficiency.

\subsection{Algorithmic Design} \label{sec:algo}

In this subsection, we discuss practical designs for applying \cref{coro:2} to image restoration.
We will adopt similar setups from prior diffusion models \citep{saharia2022palette} and assume pair information is available during training, \ie $p(X_0,X_1) = p_\calA(X_0) p_\calB(X_1|X_0)$.
From which, we can construct tractable SBs between individual data points $X_0$ and their corresponding degraded distributions $p_\calB(X_1|X_0)$.
As rebasing the terminal distribution from Gaussian to $p_\calB(\cdot|X_0)$ makes $f$ unnecessary, we will drop $f := 0$ and let \model{} learn the full nonlinear drift by itself.

\begin{algorithm}[t]
    \caption{Training}
    \label{alg:train}
    \begin{algorithmic}[1]
    \STATE {\bfseries Input:} clean $p_\calA(\cdot)$ and degraded $p_\calB(\cdot|X_0)$ datasets
    \REPEAT
    \STATE $t\sim\calU([0,1])$, $X_0 \sim p_\calA(X_0)$, $X_1 \sim p_\calB(X_1|X_0)$
    \STATE $X_t \sim q(X_t|X_0,X_1)$ according to \eqref{eq:prop3}
    \STATE Take gradient descent step on $\epsilon(X_t,t;\theta)$ using \eqref{eq:obj}
    \UNTIL{converges}
    \end{algorithmic}
 \end{algorithm}

\textbf{Sampling Proposal for Training and Generation$\quad$}
Training scalable diffusion models requires efficient computation of $X_t$.
The computation is intractable for \model{}, if directly from the nonlinear SDE \eqref{eq:fsb},  since its forward drift $\gradlog \Psi$ is not only generally nonlinear but never explicitly constructed.
Computing $X_t$ from the linear SDE \eqref{eq:fp-fsde} whose score function corresponds to $\gradlog \Psihat$ will not work either.
Since the diffusion process in \eqref{eq:fp-fsde} does \emph{not} converge to the terminal distribution (\ie $p_\calB(X_1|X_0)$) of \model{}, high-probability regions induced by \eqref{eq:fp-fsde} can be far away from regions where the generative processes actually traverse; see \cref{fig:coro2}.
We address the difficulty in the following result.
\begin{proposition}[Analytic posterior given boundary pair] \label{prop:3}
    The posterior of {\eqref{eq:sb-sde}} given some boundary pair $(X_0, X_1)$, provided $f := 0$, admits an analytic form:
        \begin{align*}
        &q(X_t|X_0, X_1) = \calN(X_t; \mu_t(X_0, X_1), \Sigma_t), \numberthis \label{eq:prop3} \\
        \mu_t &= \frac{\sigmabar_t^2}{\sigmabar_t^2 + \sigma^2_t} X_0 +
              \frac{\sigma^2_t    }{\sigmabar_t^2 + \sigma^2_t} X_1,\quad
        \Sigma_t = \frac{\sigma_t^2 \sigmabar_t^2}{\sigmabar_t^2 + \sigma^2_t} \cdot I,
    \end{align*}
    where $\sigma^2_t {:=} \int_0^t \beta_\tau \rd \tau$ and $\sigmabar^2_t {:=} \int_t^1 \beta_\tau \rd \tau$ are variances accumulated from either sides.
    Further, this posterior marginalizes the recursive posterior sampling in DDPM \eqref{eq:ddpm}:
    \begin{align*}
        q(X_n|X_0,X_N) {=} {\int} \Pi_{k=n}^{N-1} p(X_{k}|X_0, X_{k{+}1}) \rd X_{k{+}1}.
    \end{align*}
\end{proposition}

\cref{prop:3} suggests that the analytic posterior of SB given a boundary pair $(X_0, X_1)$
is the marginal density induced by DDPM, $p(X_{k}|X_0^\epsilon, X_{k{+}1})$, when $X_0^\epsilon := X_0$
and $X_N \sim p_\calB$.
Practically, this suggests that \textit{(i)} during training when $(X_0, X_1)$ are available from $p_\calA(X_0)$ and $p_\calB(X_1|X_0)$, we can sample $X_t$ directly from \eqref{eq:prop3} without solving any nonlinear diffusion as in prior SB models \citep{vargas2021solving}, and \textit{(ii)} during generation when only $X_1\sim p_\calB$ is given,
running standard DDPM starting from $X_1$ induces the same marginal density of SB paths so long as the predicted $X_0^\epsilon$ is close to $X_0$. %
Therefore, the proposed sampling proposal in \eqref{eq:prop3} is both tractable and able to cover regions traversed by generative processes.

\textbf{Parameterization \& Objective$\quad$}
Since \model{} requires no conditioning modules, we adopt the same network parameterization $\epsilon(X_t,t;\theta)$ from SGM \citep{dhariwal2021diffusion}.
Similar to the objective \eqref{eq:dsm}, we can compute the score function for $\gradlog \Psihat(X_t,t|X_0) \equiv \gradlog~p^\text{\eqref{eq:fp-fsde}}(X_t,t|X_0)$, except $X_t$ being drawn from \eqref{eq:prop3}.
This leads to
\begin{align} \label{eq:obj}
    \norm{
        \epsilon(X_t,t; \theta) - \frac{X_t {-} X_0}{\sigma_t}
    }
\end{align}
as we adopt $f:=0$.
\cref{alg:train,alg:sample} summarize the training and generation procedures of \model{}, respectively.

\begin{algorithm}[t]
    \caption{Generation}
    \label{alg:sample}
    \begin{algorithmic}[1]
    \STATE {\bfseries Input:} $X_N \sim p_\calB(X_N)$, trained $\epsilon(\cdot,\cdot; \theta)$
    \FOR{$n=N$ {\bfseries to} $1$}
    \STATE Predict $X_0^\epsilon$ using $\epsilon(X_n, t_n; \theta)$
    \STATE $X_{n-1} \sim p(X_{n-1} | X_0^\epsilon, X_n)$ according to DDPM \eqref{eq:ddpm}
    \ENDFOR
    \STATE {\bfseries return} $X_0$
    \end{algorithmic}
 \end{algorithm}

\subsection{Connection to Flow-based Optimal Transport (OT)}

It is known that the solution to SB, as an entropic optimal transport model, converges weakly to the optimal transport plan \citep{mikami2004monge}
as the diffusion degenerates.
The following result characterizes this infinitesimal limit.
\begin{proposition}[Optimal Transport ODE; OT-ODE] \label{prop:4}
    When $\beta_t \rightarrow 0$, the SDE between $(X_0,X_1)$ reduces to an ODE:
    \begin{align} \label{eq:ot}
        \rd X_t = v_t(X_t| X_0) \dt, \text{ } v_t(X_t| X_0) = \frac{\beta_t}{\sigma_t^2} (X_t - X_0),
    \end{align}
    whose solution $\mu_t(X_0,X_1)$ is the posterior mean of \eqref{eq:prop3}.
\end{proposition}
Note that the OT-ODE \eqref{eq:ot} is \emph{not} a probability flow ODE, which has the same marginal as the corresponding SDE, in the SGM literature \citep{chen2018neural,song2021maximum}. Instead, the OT-ODE \eqref{eq:ot} simulates an OT plan \citep{peyre2019computational} only when the stochasticity of the SDE vanishes.

\cref{prop:4} suggests that the mean of the posterior $q$ represents the OT-ODE paths.
Hence, \model{} can also be instantiated as a simulation-free OT by replacing the posteriors with their means, \ie by removing the noise injected into $X_t$ in both training and generation (the lines 4 in \cref{alg:train,alg:sample}).
The ratio $\frac{\beta_t}{\sigma_t^2}$ characterizes how fast the OT-ODE approaches $X_0$, in a similar vein to the noise scheduler in SGM \citep{nichol2021improved}. With this interpretation in mind, we introduce our final result, which complements recent advances in flow-matching \citep{lipman2022flow} except for image-to-image problem setups.
\begin{corollary} \label{coro:5}
    For sufficiently small $\beta_t := \beta$ that remains constant over $t$, we have $v_t = \frac{X_t - X_0}{t}$ and $\mu_t = (1-t) X_0 + t X_1$, which recover the OT displacement \citep{mccann1997convexity}.
\end{corollary}

\subsection{{\fontsize{9.5}{9.5}\selectfont Comparison to Standard Conditional Diffusion Model}}

\model{}
can be thought of as a new class of conditional diffusion models that better leverages the degraded images as the structurally informative priors.
It differs from the standard conditional SGM
(CSGM; \citet{rombach2022high,saharia2022palette}),
which simply constructs a \emph{conditional score function} with the newly available information (in this case, the degraded images) as an additional input. The generative denoising process in CSGM remains the same as the SDE \eqref{eq:rsgm} in SGM that starts from a Gaussian prior.
Intuitively, it is more efficient to learn the direct mappings between clean and degraded images given that they are already close to each other.
We summarize the comparison of I$^2$SB with other diffusion models in \cref{tab:comp_diff}.

\begin{table}[t]
    \caption{
        Comparison of different diffusion models in boundary distributions and tractability of forward and backward drifts. Note that \model{} requires pair information compared to standard SB.
    }\label{tab:comp_diff}
    \label{table:comp}
    \vskip 0.05in
    \begin{center}
    \begin{small}
    \begin{tabular}{rclcc}
    \toprule
    Model & $p(X_0)$ & $p(X_1)$ & $\gradlog \Psi$ & $\gradlog \Psihat$ \\ [0.5ex]
    \midrule
    (C)SGM
    & $p_\calA$ & $\calN(0,I)$ & 0 & tractable \\[0.5ex]
    \textbf{\model{}}
    & $p_\calA$ & $p_\calB(\cdot|X_0)$ & intractable & tractable \\[0.5ex]
    SB
    & $p_\calA$ & $p_\calB(\cdot)$ & intractable & intractable  \\
    \bottomrule
    \end{tabular}
    \end{small}
    \end{center}
    \vskip -0.15in
\end{table}

\section{Related Work} \label{sec:related-work}

\textbf{Conditional SGMs (CSGMs)} for image restoration refers to a class of diffusion models that
bias the generative processes \citep{song2020score} toward the underlying intact image of some degraded measurements. This is typically achieved by conditioning the network with the degraded images via, \eg concatenation or attention \citep{rombach2022high}.
CSGMs have demonstrated impressive results in many restoration tasks such as deblurring \citep{whang2022deblurring}, super-resolution \citep{saharia2021image}, and inpainting \citep{saharia2022palette};
yet, all of them start the generative processes from noise, which has little structural information of the clean data distribution.
\citet{pandey2022diffusevae} explored a new reparametrization of the linear forward SDE to refine a VAE's output.
In contrast, our \model{} is built on a tractable SB framework and is the first to directly bridge clean and degraded image distributions for image restoration.

\textbf{Diffusion-based inverse model (DIM)} combines inverse problem techniques \citep{song2021solving} with the diffusion priors \citep{ramesh2022hierarchical,wang2022pretraining} and aims to restore the underlying clean image signal from the (noisy) measurement given by the degraded image.
DIM typically performs projection at each generative step via, \eg Baye's rule \citep{chung2022improving,anonymous} so that the generation best aligns with the observed measurement.
This, however, requires knowing the degraded operators, whether linear \citep{kawar2022denoising,wang2022zero} or nonlinear \citep{kawar2022jpeg,chung2022diffusion}, in both training and test time.
In contrast, our \model{}, similar to other CSGMs, does not require knowing these operators, making it generally applicable without task-specific manipulations.

\section{Experiment}

\subsection{Experimental Setup}

\begin{wrapfigure}[6]{r}{0.3\columnwidth}
    \vspace{-20pt}
    \centering\includegraphics[width=0.29\columnwidth]{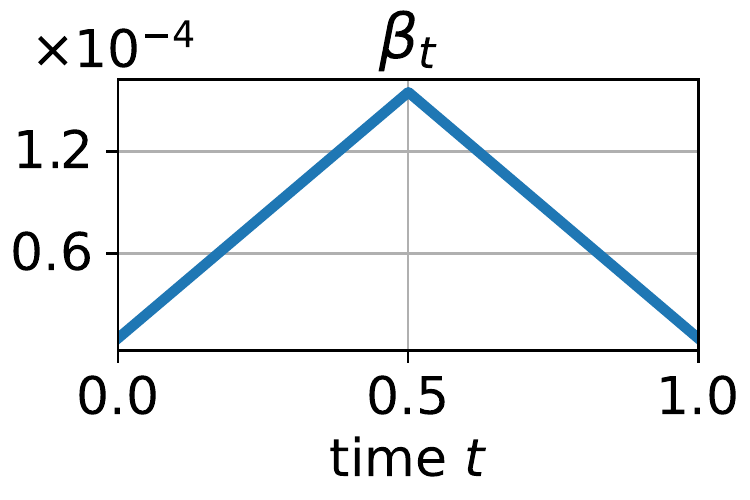}
    \vskip -0.15in
    \caption{%
        Symmetric noise scheduling.
    }
    \label{fig:noise-schedule}
  \end{wrapfigure}
\textbf{Model$\quad$}
We parameterize $\epsilon(X_t,t;\theta)$ with U-Net \citep{ronneberger2015u} and initialize the network with the unconditional ADM checkpoint \citep{dhariwal2021diffusion} trained on ImageNet 256$\times$256.
Other parameterization, \eg preconditioning \citep{karras2022elucidating}, is also applicable upon proper adaptation (see \cref{sec:edm-param}), yet we observed little performance difference.
We set $f:=0$ and consider a symmetric scheduling of $\beta_t$ where the diffusion shrinks at both boundaries; see \cref{fig:noise-schedule}.
This is suggested by prior SB models \citep{de2021diffusion,chen2021likelihood}.
By default, we use 1000 sampling time steps for all tasks with quadratic discretization \citep{song2020denoising}.

\textbf{Baselines$\quad$}
We compare \model{} with three classes of diffusion models for image restoration, namely CSGM and DIM discussed in \cref{sec:related-work} and standard SB models.
Specifically, we consider Palette \citep{saharia2022palette} and ADM \citep{dhariwal2021diffusion} for CSGM baselines.
For DIM models, we consider DDRM \citep{kawar2022denoising,kawar2022jpeg}, DDNM \citep{wang2022zero}, and $\Pi$GDM \citep{anonymous}, but stress that they require additionally knowing the corruption operators at both training and generation.
This is in contrast to CSGM models---including Palette and \model{}.
We report the results of DIM models for completeness.
Finally, for the SB baseline, we consider CDSB \citep{shi2022conditional} which extends the work of \citet{de2021diffusion} to conditional generation.

\def\SRR{SR3 \citep{saharia2021image}}
\def\Regression{Regression \citep{saharia2022palette}}
\def\ADM{ADM \citep{dhariwal2021diffusion}}
\def\Palette{Palette \citep{saharia2022palette}}
\def\DDRM{DDRM \citep{kawar2022denoising}}
\def\DDRMJPEG{DDRM \citep{kawar2022jpeg}}
\def\PGDM{$\Pi$GDM \citep{anonymous}}
\def\DDNM{DDNM \citep{wang2022zero}}
\def\CDB{{\model{} (Ours)}}
\def\CDSB{{CDSB \citep{shi2022conditional}}}

\def\CSGM{}
\def\DIRM{\rowcolor{llgray} \cellcolor{white}}

\setul{2pt}{.1pt}
\sethlcolor{llgray}

\begin{table}[t]
      \setlength\tabcolsep{5pt}
      \caption{\textbf{4$\times$ super-resolution} results w.r.t different filters.
      We report FID and Classifier Accuracy (CA, unit:\%) on a pre-trained ResNet50.
      In all \cref{table:deblur,table:sr,table:jpeg,table:inpaint},
      \hl{dark-colored rows denote methods requiring additional information such as corruption operators}, as opposed to conditional diffusion models like Palette~and our \model{}.
      }
      \label{table:sr}
      \vskip 0.08in
      \begin{center}
      \begin{small}
      \begin{tabular}{clcc}
      \toprule
      {Filter} & {Method} & {FID $\downarrow$} & {CA$\uparrow$} \\
      \midrule
      \DIRM \multirow{6}{*}{\textit{Pool}}
            & \DDRM       & 14.8 & 64.6 \\[0.5pt]
      \DIRM & \DDNM       & 9.9 & 67.1 \\[0.5pt]
       & \cellcolor{llgray}\PGDM       & \cellcolor{llgray} 3.8 & \cellcolor{llgray} \ul{72.3} \\[0.5pt]
      \CSGM & \ADM    &  \ul{3.1} & \textbf{73.4} \\[0.5pt]
      \CSGM & \CDSB   & 13.0  & 61.3 \\[0.5pt]
      \CSGM & \CDB        &  {\textbf{2.7}} & {71.0} \\
      \midrule
      \DIRM \multirow{6}{*}{\textit{Bicubic}}
            & \DDRM       & 21.3 & 63.2 \\[0.5pt]
      \DIRM & \DDNM       & 13.6 & 65.5 \\[0.5pt]
       & \cellcolor{llgray}\PGDM       & \cellcolor{llgray} \ul{3.6} & \cellcolor{llgray} \textbf{72.1} \\[0.5pt]
      \CSGM & \ADM    & 14.8 & 66.7 \\[0.5pt]
      \CSGM & \CDSB   &  13.6 & 61.0 \\[0.5pt]
      \CSGM & \CDB        &  {\textbf{2.8}} & {\ul{70.7}} \\
      \bottomrule
      \end{tabular}
      \end{small}
      \end{center}
      \vskip -0.22in
  \end{table}

\begin{table}[t]
    \caption{\textbf{JPEG restoration} w.r.t different quality factors (QF).
    }
    \label{table:jpeg}
    \vskip 0.07in
    \begin{center}
    \begin{small}
    \begin{tabular}{clcc}
    \toprule
    {QF} & {Method} & {FID-10k $\downarrow$} & {CA$\uparrow$} \\
    \midrule
    \DIRM %
          & \DDRMJPEG   & 28.2 & 53.9 \\[0.5pt]
    \DIRM & \PGDM       &  8.6 & 64.1 \\[0.5pt]
    \CSGM 5 & \Palette    &  \ul{8.3} & \ul{64.2} \\[0.5pt]
    \CSGM & \CDSB   & 38.7  & 45.7 \\[0.5pt]
    \CSGM & \CDB        &  {\textbf{4.6}} & {\textbf{67.9}} \\
    \midrule
    \DIRM %
          & \DDRMJPEG   & 16.7 & 64.7 \\[0.5pt]
    \DIRM & \PGDM       &  6.0 & \ul{71.0} \\[0.5pt]
    \CSGM 10 & \Palette    &  \ul{5.4} & 70.7 \\[0.5pt]
    \CSGM & \CDSB   & 18.6  & 60.0 \\[0.5pt]
    \CSGM & \CDB        &  \textbf{3.6} & \textbf{72.1} \\
    \bottomrule
    \end{tabular}
    \end{small}
    \end{center}
    \vskip -0.1in
\end{table}

\textbf{Evaluation$\quad$}
We showcase the performance of \model{} in solving various image restoration problems, including inpainting, JPEG restoration, deblurring, and 4$\times$ super-resolution (64$\times$64 to 256$\times$256), on ImageNet 256$\times$256.
For each restoration problem, we consider 2-3 tasks by varying, \eg the quality factors, filtering kernels, and mask types.
We keep the implementation and setup of each restoration task the same as the baselines \citep{kawar2022denoising,kawar2022jpeg,saharia2022palette} for a fair comparison; see \cref{sec:exp-detail} for details.
For quantitative metrics, we choose
the Frechet Inception Distance (FID; \citet{heusel2017gans}) and Classifier Accuracy (CA) of a pre-trained ResNet50 \citep{he2016deep}.
Similar to the baselines \citep{saharia2022palette,anonymous}, we report super-resolution results on the full ImageNet validation set and report the remaining results on a 10k validation subset.\footnote{
 \url{https://bit.ly/eval-pix2pix}
}

\def\SRR{SR3 \citep{saharia2021image}}
\def\Regression{Regression \citep{saharia2022palette}}
\def\Palette{Palette \citep{saharia2022palette}}
\def\DDRM{DDRM \citep{kawar2022denoising}}
\def\PGDM{$\Pi$GDM \citep{anonymous}}
\def\DDNM{DDNM \citep{wang2022zero}}
\def\CDB{{\model{} (Ours)}}
\def\CDSB{{CDSB \citep{shi2022conditional}}}

\def\CSGM{}
\def\DIRM{\rowcolor{llgray} \cellcolor{white}}

\setul{2pt}{.1pt}
\sethlcolor{llgray}

\begin{table}[t]
    \setlength\tabcolsep{4.5pt}
    \caption{\textbf{Inpainting} results w.r.t different masks.}
    \label{table:inpaint}
    \vskip -0.05in
    \begin{center}
    \begin{small}
    \begin{tabular}{clcc}
    \toprule
    {Mask} & {Method} & {FID-10k $\downarrow$} & {CA$\uparrow$} \\
    \midrule
    \DIRM %
     & \DDRM    & 24.4 & 62.1 \\[0.5pt]
    \DIRM & \PGDM    & 7.3 & \textbf{72.6} \\[0.5pt]
    \textit{Center} & \cellcolor{llgray}\DDNM    & \cellcolor{llgray}15.1 & \cellcolor{llgray}55.9 \\[0.5pt]
    \CSGM \textit{128$\times$128} & \Palette &  \ul{6.1} & {63.0} \\[0.5pt]
    \CSGM & \CDSB   & 50.5  & 49.6 \\[0.5pt]
    \CSGM & \CDB  &  {\textbf{4.9}} & {\ul{66.1}} \\
    \midrule
    \DIRM \multirow{5}{*}{\specialcell{ \cellcolor{white} \textit{Freeform} \\ \cellcolor{white} \textit{10\%-20\%}}}
        & \DDRM    & 9.7 & 67.6 \\[0.5pt]
     & \cellcolor{llgray}\DDNM    & \cellcolor{llgray} \ul{3.2} & \cellcolor{llgray} 73.6 \\[0.5pt]
    \CSGM & \Palette & 4.0 & \ul{73.7} \\[0.5pt]
    \CSGM & \CDSB   &  8.5 & 71.2 \\[0.5pt]
    \CSGM & \CDB  & \textbf{2.9} & \textbf{74.9} \\
    \midrule
    \DIRM %
    & \DDRM    & 8.6 & 71.9 \\[0.5pt]
    \DIRM & \PGDM    & 5.3 & \textbf{75.3} \\[0.5pt]
    \textit{Freeform} & \cellcolor{llgray}\DDNM    & \cellcolor{llgray}\ul{4.2} & \cellcolor{llgray}70.8 \\[0.5pt]
    \CSGM \textit{20\%-30\%} & \Palette &  4.1 & 71.8 \\[0.5pt]
    \CSGM & \CDSB   & 16.5 & 64.5 \\[0.5pt]
    \CSGM & \CDB  & \textbf{3.2} & \ul{73.4} \\
    \bottomrule
    \end{tabular}
    \end{small}
    \end{center}
    \vskip -0.2in
\end{table}

\begin{table}[t]
    \setlength\tabcolsep{4.5pt}
    \caption{\textbf{Deblurring} results w.r.t different kernels.
    }
    \label{table:deblur}
    \vskip 0.07in
    \begin{center}
    \begin{small}
    \begin{tabular}{clcc}
    \toprule
    {Kernel} & {Method} & {FID-10k $\downarrow$} & {CA$\uparrow$} \\
    \midrule
    \DIRM %
          & \DDRM       & 9.9 & 68.0 \\[0.5pt]
    \DIRM & \DDNM & \textbf{3.0} & \textbf{75.5} \\[0.5pt]
    \CSGM \textit{Uniform} & \Palette    & 4.1 & \ul{74.0} \\[0.5pt]
    \CSGM & \CDSB   & 15.5 & 65.1 \\[0.5pt]
    \CSGM & \CDB        & {\ul{3.9}} & {73.7} \\
    \midrule
    \DIRM %
          & \DDRM       & 6.1 & 72.5 \\[0.5pt]
    & \cellcolor{llgray}\DDNM & \cellcolor{llgray}\textbf{2.9} & \cellcolor{llgray}\textbf{75.6} \\[0.5pt]
    \CSGM \textit{Gaussian} & \Palette    & 3.1 & \ul{75.4} \\[0.5pt]
    \CSGM & \CDSB   & 7.7 & 71.1 \\[0.5pt]
    \CSGM & \CDB        & \ul{3.0} & 75.0 \\
    \bottomrule
    \end{tabular}
    \end{small}
    \end{center}
    \vskip -0.15in
\end{table}

\subsection{Experimental Results}

\textbf{\model{} surpasses standard CSGM on many tasks$\quad$}
\cref{table:sr,table:jpeg,table:inpaint,table:deblur} summarize the quantitative results on each restoration task.
We use the official values reported by each baseline and, if not available, compute them using the official implementations with default hyperparameters,
except for Palette on deblurring and inpainting tasks which we implemented by ourselves.
\model{} clearly surpasses standard CSGMs such as Palette and ADM on six out of nine tasks, including super-resolution (\textit{Bicubic}), JPEG restoration (for both QFs), and inpainting (for all masks).
Despite that ADM and Palette obtain higher CA on super-resolution (\textit{Pool}) and both deblurring tasks, \model{} yields lower, hence better, FID.

\textbf{\model{} matches DIM without knowing corrupted operators and outperforms standard SB on all tasks$\quad$}
Compared to \hl{DIM models}, \model{} provides a competitive alternative with similar performance yet \emph{without} knowing the corrupted operators during either training or generation.
In fact, \model{} achieves state-of-the-art FID on seven out of nine tasks and set new records for CA on JPEG restoration (both QFs) and inpainting (\textit{Freeform 10-20\%}).
Finally, \model{} outperforms CDSB on \emph{all} restoration tasks by a large margin.
These results highlight \model{} as the \emph{first} nonlinear diffusion model that scales to high-dimensional applications.

\begin{figure}[t]
    \vskip 0.05in
    \centering
    \begin{minipage}{0.48\textwidth}
        \begin{center}
            \centerline{\includegraphics[width=\columnwidth]{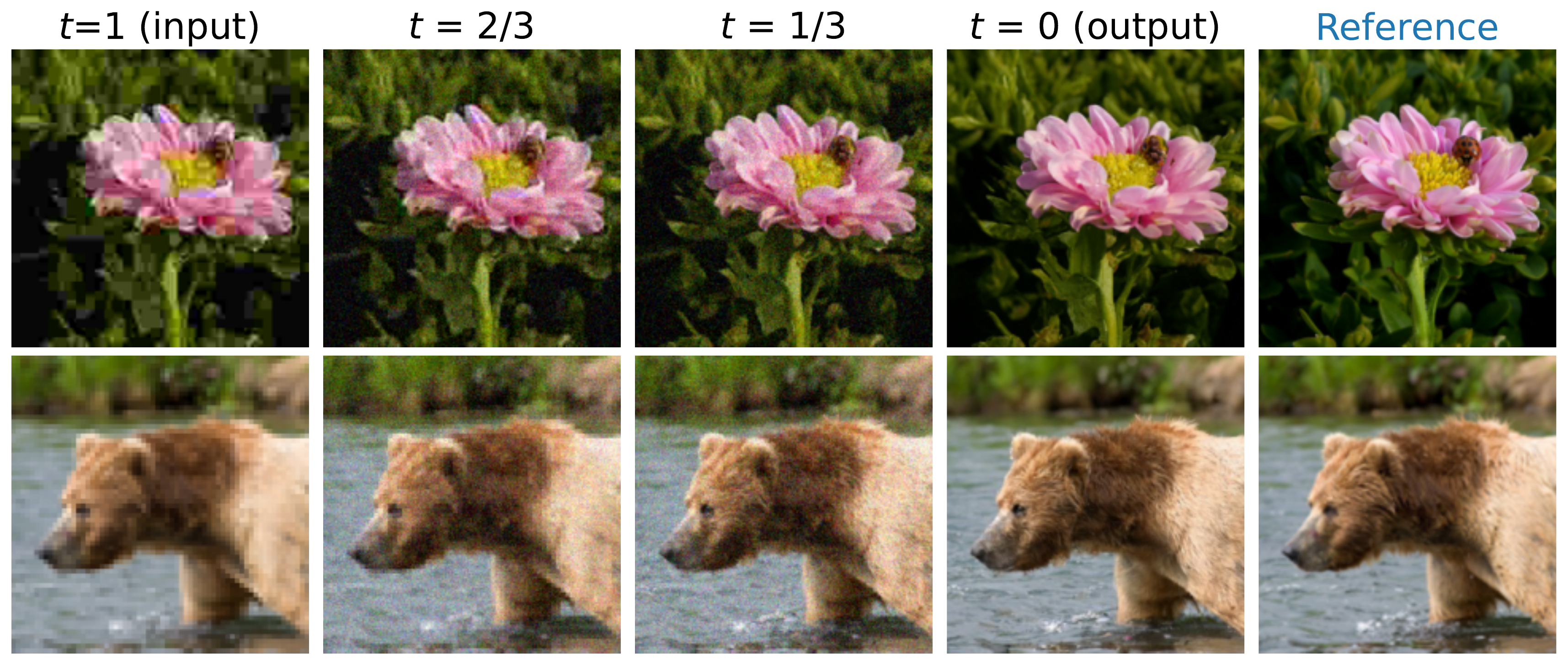}}
            \vskip -0.15in
            \caption{
                \model{} features more natural and interpretable generative diffusion processes from degraded to clean images.
                \textbf{Top}: JPEG restoration (QF=5).
                \textbf{Bottom}: 4$\times$ super-resolution (\textit{Bicubic}).
            }
            \label{fig:generation}
        \end{center}
    \end{minipage}
    \vskip 0.05in
    \begin{minipage}{0.48\textwidth}
        \begin{center}
            \centerline{\includegraphics[width=0.95\columnwidth]{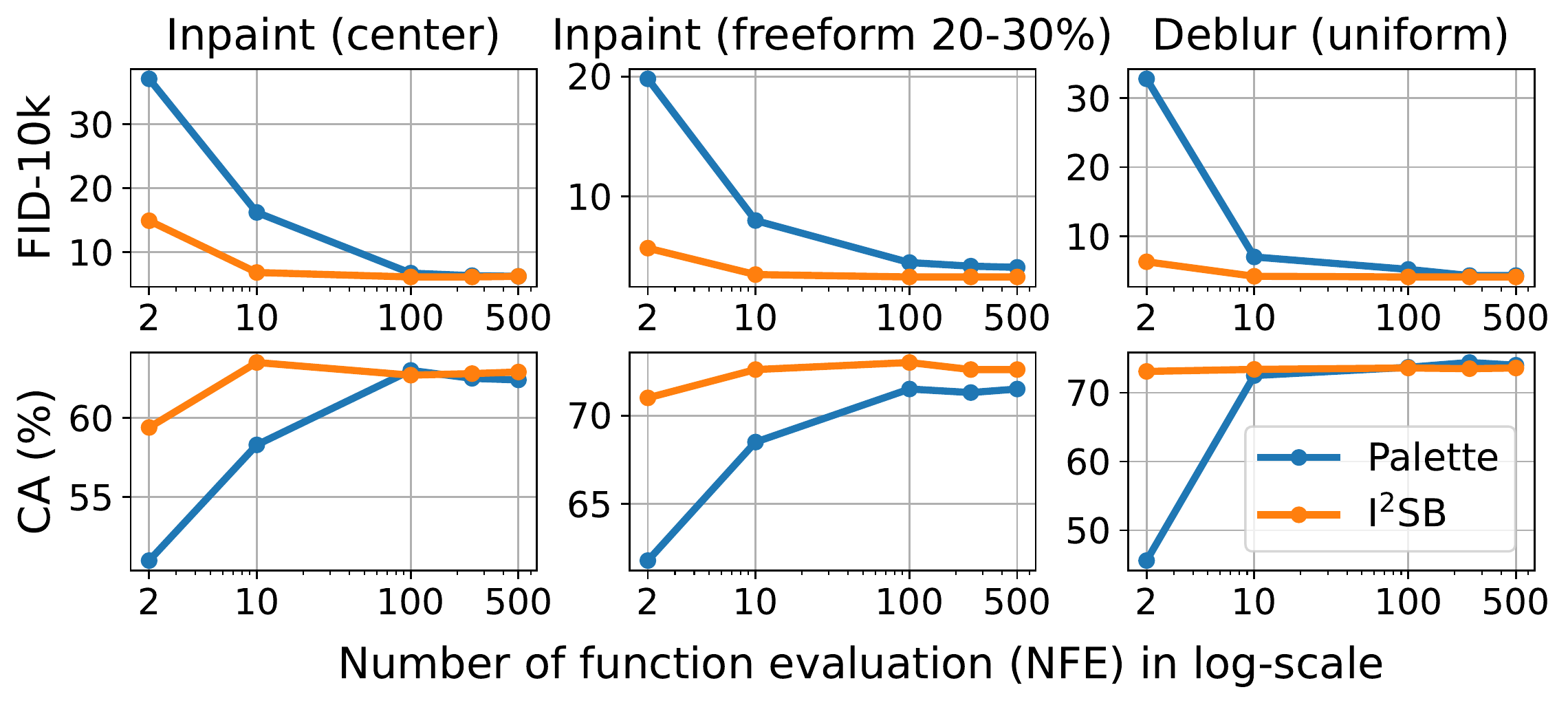}}
            \vskip -0.15in
            \caption{
                Quantitative comparison between Palette \citep{saharia2022palette} and our \model{} across different NFEs in sampling. \model{} enjoys much smaller performance drops as NFE decreases.
            }
            \label{fig:nfe-curve}
        \end{center}
    \end{minipage}
    \vskip -0.15in
\end{figure}

\begin{table}[t]
    \vskip -0.1in
    \caption{
        How the performance of \model{} \markgreen{improves} or \markred{degrades} with OT-ODE, \ie by sampling $X_t$ from the mean of $q(X_t|X_0,X_1)$.
    }
    \label{table:aba-ot}
    \begin{center}
    \begin{small}
    \begin{tabular}{lcccc}
        \toprule
        & \multicolumn{2}{c}{JPEG restoration} & \multicolumn{2}{c}{Deblurring} \\[0.5pt]
        & QF=5 & 10 & \textit{Uniform} & \textit{Gaussian} \\
        \midrule
        FID difference & \markred{+5.3} & \markred{+4.2} & \markgreen{-0.3} & \markgreen{-0.6} \\[0.5pt]
        CA  difference & \markred{-4.7} & \markred{-3.8} & \markgreen{+6.0} & \markgreen{+4.1} \\
        \bottomrule
        \end{tabular}
    \end{small}
    \end{center}
    \vskip -0.1in
\end{table}

\begin{figure}[t]
    \vskip 0.05in
    \centering
    \begin{minipage}{0.48\textwidth}
        \begin{center}
            \centerline{\includegraphics[width=\columnwidth]{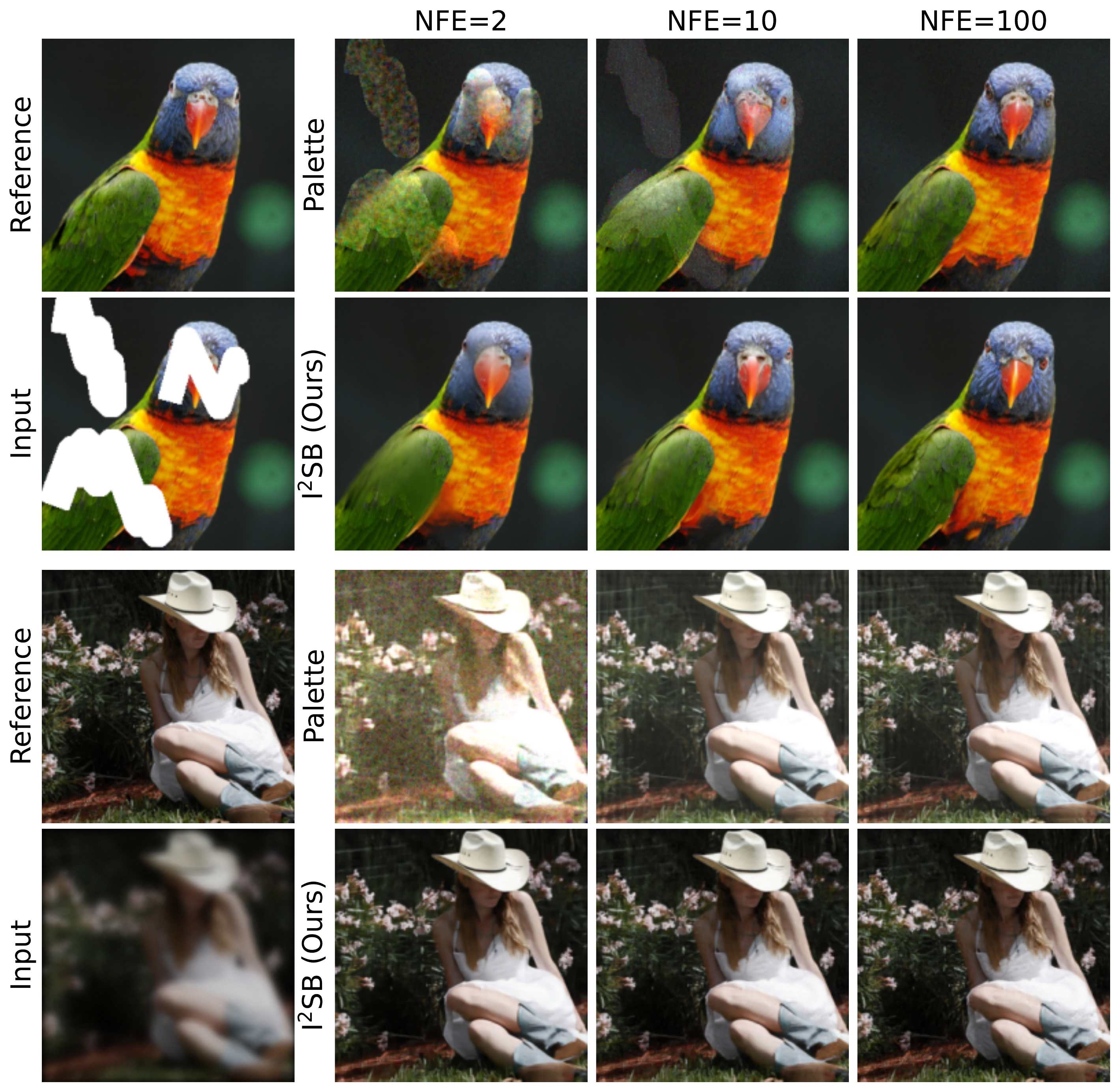}}
            \vskip -0.15in
            \caption{
                Qualitative comparison between Palette \citep{saharia2022palette} and our \model{} w.r.t. different NFE on
                (\textbf{top}) inpainting (\textit{Freeform 20\%-30\%})
                and (\textbf{bottom}) deblurring (\textit{Uniform}).
            }
            \label{fig:nfe-visual}
        \end{center}
    \end{minipage}
    \begin{minipage}{0.48\textwidth}
        \begin{center}
            \centerline{\includegraphics[width=0.77\columnwidth]{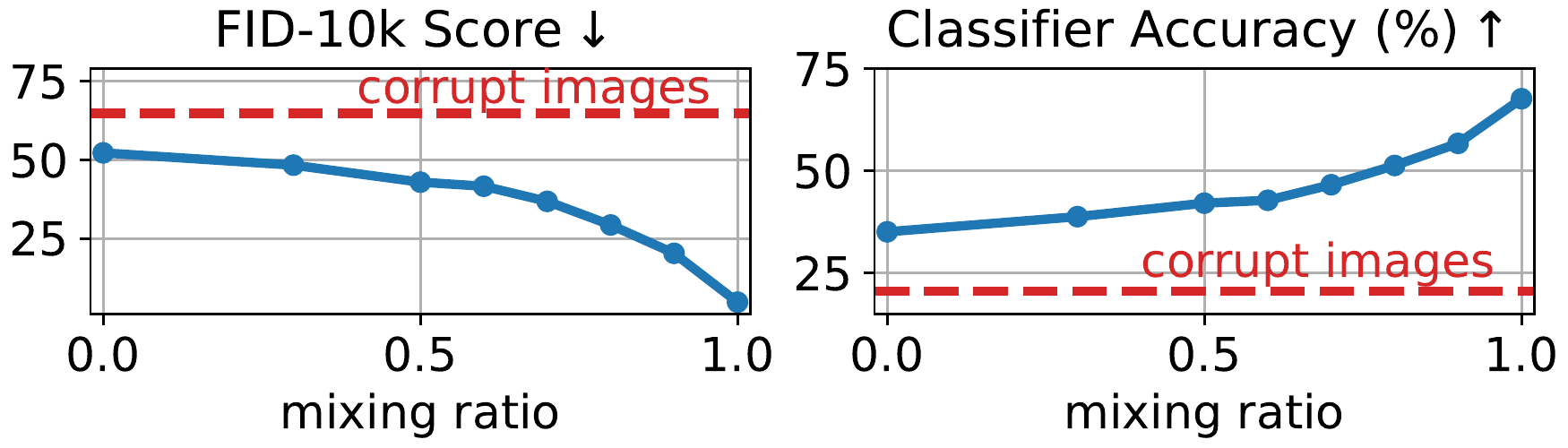}}
            \vskip -0.12in
            \caption{
                Effect of sampling proposal of $X_t$ on JPEG restoration (QF=5).
                The $x$-axis is the mixing ratio between (\textbf{left}) the distribution induced by \eqref{eq:fp-fsde} and (\textbf{right}) the posterior $q(X_t|X_0,X_1)$.
                Both metrics improve as the proposal approaches $q(X_t|X_0,X_1)$.
            }
            \label{fig:aba-samp}
            \end{center}
    \end{minipage}
    \vskip -0.15in
\end{figure}

\textbf{\model{} yields interpretable \& efficient generation$\quad$}
As \model{} directly constructs diffusion bridges between two domains,
it generates more interpretable processes that progressively {restore the intact images from the degradations}; see \cref{fig:generation}.
\emph{More interpretable generation also implies sampling efficiency}.
Since the clean and degraded images are typically close to each other,
the generation of \model{} starts from a much more structurally informative prior compared to random noise.
We validate these concepts in \cref{fig:nfe-visual,fig:nfe-curve} by tracking how the performance of \model{} and Palette changes as the number of function evaluation (NFE) decreases in sampling.
For a fair comparison, we train both models with 1000 discrete steps and sample with DDPM \eqref{eq:ddpm} so that they differ mainly in the boundary distributions, \ie $p_\calB(\cdot|X_0)$ \textit{vs.} $\calN(0,I)$.
From \cref{fig:nfe-curve}, we see that across various tasks, \model{} enjoys much smaller performance drops as NFE decreases.
On inpainting (\textit{Freeform 20-30\%}), for example, \model{} needs only 2$\sim$10 NFEs while Palette needs at least 100 NFEs to achieve the similar best performance.
Qualitatively, \cref{fig:nfe-visual} also demonstrates that \model{} clearly outperforms Palette in the small NFE regime. Particularly for inpainting, \model{}
is able to repaint the masked region with semantic structures with only two NFEs (and further fills in textural details as the NFE increases).
On the contrary, Palette tends to generate unnatural images with noisy repainting or contrast shift when the NFE is small.

\begin{figure*}
    \vskip 0.05in
    \begin{center}
    \centerline{\includegraphics[width=\textwidth]{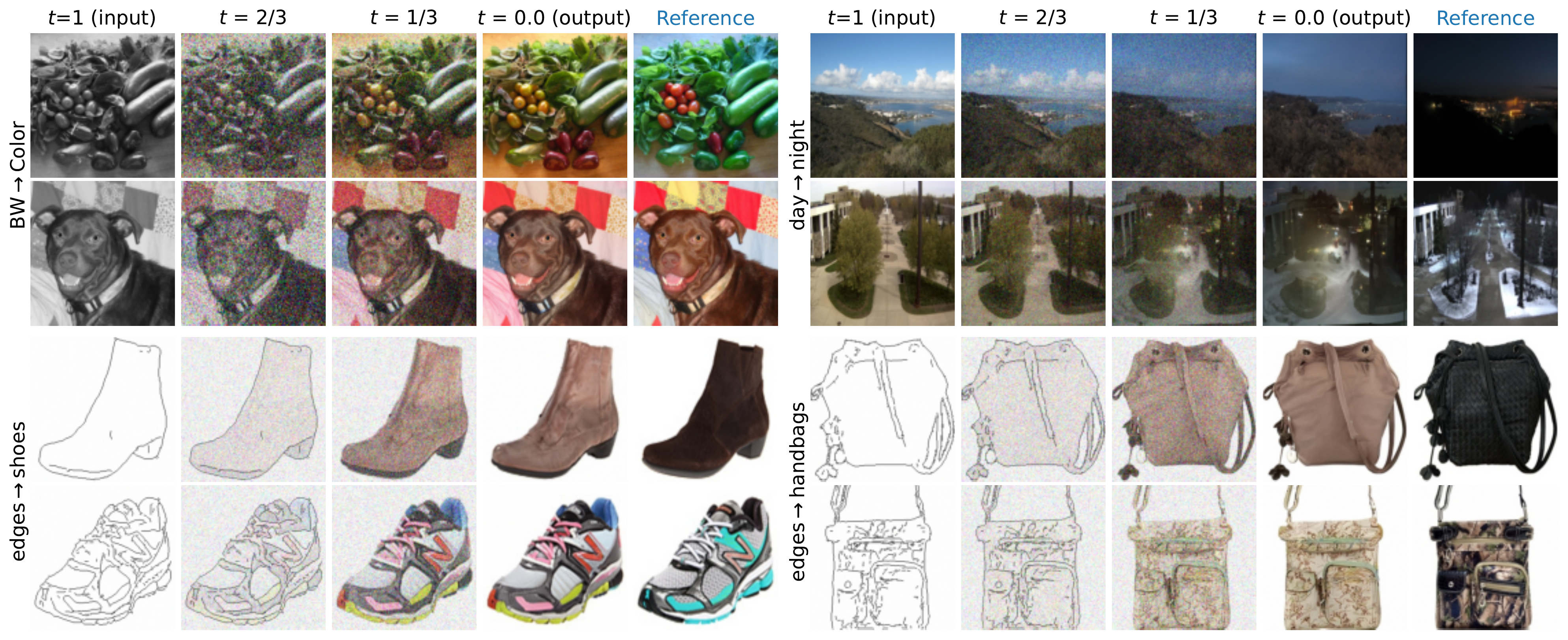}}
    \vskip -0.1in
    \caption{
        Application of our \model{} to four \textbf{general image-to-image translation} tasks from Pix2pix \citep{isola2017image}.
        We consider ImageNet dataset for the colorization task (BW $\rightarrow$ Color) and adopt the datasets proposed by \citet{isola2017image} for the remaining three tasks, namely \texttt{edges2shoes}, \texttt{day2night}, and \texttt{edges2handbags}. All images are in 256$\times$256 resolution.
    }
    \label{fig:pix2pix}
    \end{center}
    \vskip -0.2in
\end{figure*}

\subsection{Discussions}

\textbf{Sampling proposals$\quad$}
\model{} shares much algorithmic similarity with SGM except drawing $X_t$
from an interpolation between clean and degraded images according to $q(X_t|X_0,X_1)$.
This posterior differs from the distribution induced by the forward SDE \eqref{eq:fp-fsde} and, according to \cref{prop:3}, better covers regions traversed by the generative processes.
To verify this, \cref{fig:aba-samp} shows how the performance changes when $X_t$ is sampled by mixing these two distributions with different ratios during training.
Clearly, both metrics deteriorate as the sampling proposal deviates from $q(X_t|X_0,X_1)$ towards the distribution induced by \eqref{eq:fp-fsde}.

\textbf{Diffusion \textit{vs.} OT-ODE$\quad$}
\cref{table:aba-ot} reports the performance difference when we adopt the OT-ODE in \cref{prop:4}, \ie by sampling $X_t$ with the mean of $q(X_t|X_0,X_1)$ in both training and generation.
Our result suggests that OT-ODE favors restoration tasks where deterministic mapping is possible (\eg deblurring) yet is biased against those with large uncertainties (\eg JPEG restoration).
It reexamines the role of stochasticity in modern dynamic generative models.

\begin{table}[t]
    \vskip -0.2in
    \setlength\tabcolsep{4.5pt}
    \caption{
        Quantitative (FID) results on two general image-to-image translation tasks. Our \model{} matches Pix2pix with only \emph{one} NFE and quickly outperforms it by refining the generation processes.
    }
    \label{table:pix2pix}
    \begin{center}
    \begin{small}
    \begin{tabular}{lcccc}
        \toprule
        & \multirow{2}{*}{Pix2pix} & \multicolumn{3}{c}{\model{}} \\[0.5pt]
         & & NFE=1 & NFE=5 & NFE=1000 \\
        \midrule
        edges$\rightarrow$shoes & 73.9 & 73.9 & 54.2 & \textbf{37.8} \\[0.5pt]
        day$\rightarrow$night & 196.4 & 196.3 & 185.8 & \textbf{153.6} \\
        \bottomrule
        \end{tabular}
    \end{small}
    \end{center}
    \vskip -0.1in
\end{table}

\textbf{General image-to-image translation$\quad$}
Since our framework does not impose any assumptions or restrictions on the underlying prior distributions, \model{} can be applied to general image-to-image translation by adopting the same training and sampling procedures (\cref{alg:train,alg:sample}), except conditioning the network additionally on the inputs, \ie $\epsilon(X_t,t,X_1|\theta)$.
Aligned with the discussions in \cref{sec:more-discuss}, we found it beneficial when the priors have large information loss.
\cref{fig:pix2pix} demonstrates the qualitative results, and \cref{table:pix2pix} reports the FID w.r.t. the statistics of each validation set.
It is clear that our \model{} achieves similar performance to Pix2pix \citep{isola2017image} with \emph{one} NFE and quickly outperforms it by refining the generation processes. These results highlight the applicability of \model{} to general image-to-image translation tasks.

\textbf{Comparison to inpainting GANs$\quad$}
\cref{table:gan} reports the generation quality and efficiency between two inpainting GANs, \ie DeepFillv2 \citep{yu2019free} and HiFill \citep{yi2020contextual}, Palette, and our \model{}.
For a fair comparison, we reduce the sampling step of all diffusion models to 1. In other words, ``{\model{} (NFE=1)}'' generates images with \emph{one} network call.
It is clear that \model{} achieves best generation quality among all models on both tasks.
Note that since all models generate images in one network call, the difference in their inference times is mainly due to the network size.

\begin{table}[t]
    \vskip -0.2in
    \setlength\tabcolsep{4.5pt}
    \caption{Comparison between GANs (DFill and HiFill) and our \model{} with \emph{one} NFE on two inpainting tasks. We include Palette for comparison. The inference time is measured on a V100 16G.}
    \vskip 0.01in
    \label{table:gan}
    \begin{center}
    \begin{small}
    \begin{tabular}{clccc}
    \toprule
    \multirow{2}{*}{Mask} & \multirow{2}{*}{Method} & FID $\downarrow$ & CA $\uparrow$ & {Inference time} \\
     &  & (10k) & (\%) & {(sec/image)} \\
    \midrule
    & DeepFillv2 & 6.7 & 71.6 & \textbf{0.01} \\[0.5pt]
    \textit{Freeform} & HiFill & 7.5  & 70.1 & 0.03\\[0.5pt]
    \textit{10\%-20\%} & \model{} (NFE=1) & \textbf{4.1} & \textbf{73.4} & 0.14 \\[0.5pt]
    & Palette (NFE=1) & 9.6 & 69.9 & 0.14 \\
    \midrule
    & DeepFillv2 & 9.4 & 68.8 & \textbf{0.01} \\[0.5pt]
    \textit{Freeform} & HiFill & 12.4 & 65.7 & 0.03\\[0.5pt]
    \textit{20\%-30\%} & \model{} (NFE=1) & \textbf{6.7} & \textbf{69.9} & 0.14 \\[0.5pt]
    & Palette (NFE=1) & 19.8 & 61.8 & 0.14 \\
    \bottomrule
    \end{tabular}
    \end{small}
    \end{center}
    \vskip -0.1in
\end{table}

\textbf{Limitation$\quad$}
Despite these encouraging results, the tractability of \model{} requires knowing paired data (\eg clean and degraded image pairs) during training. While paired data is typically available at nearly no cost, especially for image restoration tasks, it nevertheless limits the application of \model{} to \emph{unpaired} image translation tasks like
CycleGAN \citep{zhu2017unpaired} or DDIB \citep{su2022dual}.
Constructing simulation-free diffusion bridges (like our \model{}) under more flexible setups will be an interesting future direction.

\section{Conclusion}
We developed \model{}, a new conditional diffusion model that transport between clean and degraded image distributions based on a tractable class of Schr\"odinger bridge.
\model{} yields interpretable generation, enjoys sampling efficiency, and sets new records on image restoration.
It will be interesting to combine \model{} with inverse problem techniques.

\vspace{-5pt}
\section*{Acknowledgements}
\vspace{-5pt}
The authors thank Jiaming Song \& Yinhuai Wang for experiment clarifications, Jeffrey Smith \& Sabu Nadarajan for hardware supports, Tianrong Chen for general discussions, and David Zhang for catching typos in the initial arXiv.

\bibliography{reference}
\bibliographystyle{icml2023}

\clearpage
\appendix

\section{Proof} \label{sec:proof}

\begin{proof}[\textbf{Proof of \cref{thm:1}}]
    Recall that the density evolution of an It{\^o} process,
    \begin{align} \label{eq:p1}
        \rd X_t = f_t(X_t) \dt + \sbeta \rd W_t, \quad X_0 \sim p_0
    \end{align}
    can be characterized by the Fokker Plank equation \citep{risken1996fokker},
    \begin{align} \label{eq:p2}
        \fracpartial{p(x,t)}{t} {=} - \nabla \cdot (f_t~p) + \frac{1}{2} \beta_t \Delta p,
         p(x,0) = p_0(x).
    \end{align}
    Comparing (\ref{eq:p1}, \ref{eq:p2}) to  (\ref{eq:fp-fsde}, \ref{eq:sb-pde}) readily suggests that the PDE $\fracpartial{\Psihat(x,t)}{t}$ in \eqref{eq:sb-pde} can be viewed as the Fokker Plank of the SDE in \eqref{eq:fp-fsde}. The equivalence $\Psihat \equiv p^\text{\eqref{eq:fp-fsde}}$ holds up to some constant which vanishes upon taking the operator ``$\gradlog$'' or in the Fokker Plank equation (since all operators are linear). Similar interpretation can be drawn between the PDE $\fracpartial{\Psi(x,t)}{t}$ and the SDE in \eqref{eq:fp-rsde} by noticing that \eqref{eq:sb-pde} can be read \emph{equivalently} from the reversed time coordinate \citep{chen2021likelihood,liu2022deep}:
    \begin{align}
        \numberthis \label{eq:p3}
        \begin{cases}
        \fracpartial{\Psi(x,s)}{s}    = \nabla \cdot (\Psihat f_s) + \frac{1}{2} \beta_s \Delta \Psi \\[3pt]
        \fracpartial{\Psihat(x,s)}{s} = \nabla \Psi^\T f_s - \frac{1}{2} \beta_s \Delta \Psihat
        \end{cases},
    \end{align}
    where $s:= 1-t$. This suggests that $\Psi(x,s)$ can be seen as the density (up to some constant) of the SDE %
    \begin{align*}
        \rd X_s = - f_s(X_s) \ds + \sqrt{\beta_s} \rd W_s, \quad X_0 \sim \Psi(\cdot,0),
    \end{align*}
    which equals \eqref{eq:fp-rsde} after substituting back $t = 1-s$. %
\end{proof}

\begin{proof}[\textbf{Proof of \cref{coro:2}}]
    It suffices to show that the solutions \eqref{eq:coro2} are consistent with the necessary conditions in \eqref{eq:sb-pde}, \ie they are the solutions to the two PDEs with the coupled boundary constraints.
    Notice that the second PDE $\fracpartial{\Psihat(x,t)}{t}$ and the constraint $\Psi(\cdot,1) \Psihat(\cdot,1)=p_\calB(x)$ are satisfied by construction since $\Psihat(\cdot, 1)$ is the Fokker-Plank solution w.r.t. the initial condition $\Psihat(\cdot, 0) = \delta_{a}(\cdot)$.
    Hence, it remains to be shown that the solution to the following \emph{backward} PDE
    \begin{align} \label{eq:p4}
        \fracpartial{\Psi(x,t)}{t} = - \nabla \Psi^\T f_t - \frac{1}{2} \beta_t \Delta \Psi, \text{ }\text{ } \Psi(x,1) = \frac{p_\calB(x)}{\Psihat(x, 1)}
    \end{align}
    satisfies the remaining boundary constraint w.r.t. $p_\calA$.
    Precisely, since $p_\calA(x) = \Psihat(x, 0) = \delta_{a}(x)$, it suffices to show the solution to \eqref{eq:p4} being $\Psi(a,0) = 1$, which is indeed the case \citep{zhang2021path}.
    For completeness, \citet[Theorem 1]{zhang2021path} identified that
    the solution to the Hamilton-Jacobi-Bellman (HJB) equation \citep{evans2010partial}, which relates to \eqref{eq:p4} via exponential transform \citep{hopf1950partial,caluya2021wasserstein}, with the terminal cost $\log \frac{p_\calB(x)}{\Psihat(x, 1)}$ is simply $0$.
    Hence, we know that the solution to \eqref{eq:p4} is $\Psi(a,0) = \exp(0) = 1$, which concludes the proof.
\end{proof}

\begin{proof}[\textbf{How \cref{coro:2} reduces to SGM}]
    When $p_\calB := \Psihat(\cdot, 1)$ and $f$ is chosen such that the terminal distribution of the forward SDE converges to a Gaussian, \ie $\Psihat(\cdot, 1) \approx \calN(0,I)$, we have $\Psi(\cdot,1)=1$ from \eqref{eq:coro2}. In fact, we will have $\Psi(\cdot,t)=1$ for all $t\in[0,1]$ since $\fracpartial{\Psi(x,t)}{t}= 0$. In this case, one can verify that the remaining boundary constraint holds, \ie $p_\calA(\cdot) = \Psi(\cdot,0)\Psihat(\cdot,0)$, since we set $p_\calA(\cdot) = \Psihat(\cdot,0) = \delta_a(\cdot)$.
\end{proof}

\def\denomt{{ \sigmabar_{n}^2 + \sigma^2_{n} }}
\def\denomtnxt{{ \sigmabar_{n{+}1}^2 + \sigma^2_{n{+}1} }}

\begin{proof}[\textbf{Proof of \cref{prop:3}}]
    \cref{eq:prop3} arises naturally by first conditioning the Nelson's duality \citep{nelson2020dynamical}, \ie $q(\cdot, t) = {\Psi}(\cdot, t) {\Psihat}(\cdot, t)$, on a boundary pair $(X_0,X_1)$,
    \begin{align*}
        q(X_t|X_0,X_1) = {\Psi}(X_t,t | X_0) {\Psihat}(X_t,t|X_1).
    \end{align*}
    Since $\Psi(X_t,t | X_0)$ and $\Psihat(X_t,t|X_1)$ are solutions to Fokker-Plank equations (see the proof of \cref{thm:1}), we can rewrite the posterior as the product of two Gaussians:
    \begin{align*}
        &{\Psi}(X_t,t | X_0) {\Psihat}(X_t,t|X_1) \\
        = &\exp\pr{ -\frac{1}{2} \pr{\frac{\norm{X_t-X_0}^2}{\sigma_t^2} + \frac{\norm{X_t-X_1}^2}{\sigmabar_t^2} }} \\
        = &\calN(X_t; \frac{\sigmabar_t^2}{\sigmabar_t^2 + \sigma^2_t} X_0 +
        \frac{\sigma^2_t    }{\sigmabar_t^2 + \sigma^2_t} X_1, \frac{\sigma_t^2 \sigmabar_t^2}{\sigmabar_t^2 + \sigma^2_t} \cdot I),
    \end{align*}
    where $\sigma^2_t := \int_0^t \beta_\tau \rd \tau$ and $\sigmabar^2_t := \int_t^1 \beta_\tau \rd \tau$ are analytic marginal variances \citep{sarkka2019applied} of the SDEs \eqref{eq:fp-sde} when $f:=0$.

    We now prove (by induction) that $q(X_t|X_0,X_1)$ is the marginal density of DDPM posterior $p(X_n|X_0,X_{n+1})$. First, notice that when $f:=0$, $p(X_n|X_0,X_{n+1})$ has an analytic Gaussian form
    \begin{align*}
        &p(X_n|X_0,X_{n+1}) \\
        =&~\calN(X_n; \frac{\alpha_n^2}{\alpha_n^2 + \sigma^2_n} X_0 +
        \frac{\sigma^2_n    }{\alpha_n^2 + \sigma^2_n} X_{n+1}, \frac{\sigma_n^2 \alpha_n^2}{\alpha_n^2 + \sigma^2_n} \cdot I),
    \end{align*}
    where we denote $\alpha_n^2 := \int_{t_{n}}^{t_{n+1}} \beta_\tau \rd \tau$ as the accumulated variance between two consecutive time steps $(t_n,t_{n+1})$.
    It is clear that at the boundary $t_n:=t_{N-1}$, we have
    \begin{align*}
        q(X_{N-1}|X_0,X_N) = p(X_{N-1}|X_0,X_N)
    \end{align*}
    since $\alpha_{N-1} = \int_{t_{N-1}}^{t_{N}} \beta_\tau \rd \tau = \sigmabar_{N-1}^2$.
    Suppose the relation also holds at $t_{n+1}$, it suffices to show that
    \begin{align*}
        &q(X_n|X_0,X_N) \numberthis \label{eq:p5} \\
        \overset{\text{?}}{=} &\int p (X_n|X_0,X_{n+1}) q (X_{n+1}|X_0, X_N) \rd X_{n+1}.
    \end{align*}
    Since both $p$ and $q$ are Gaussians, the RHS of \eqref{eq:p5} is a Gaussian with the mean \citep{bishop2006pattern}
    \begin{align*}
        & \frac{\alpha_n^2}{\underbrace{\alpha_n^2 {+} \sigma^2_n}_{\sigma^2_{n{+}1}}} X_0 {+} \frac{\sigma_n^2}{\underbrace{\alpha_n^2 {+} \sigma^2_n}_{\sigma^2_{n{+}1}}} \biggl(
          \frac{\sigmabar_{n{+}1}^2~X_0 }{\underbrace{\denomtnxt}_{\denomt}} +
          \frac{\sigma^2_{n{+}1}~X_N }{\underbrace{\denomtnxt}_{\denomt}}
        \biggr) \\
        =& \frac{\alpha_n^2 (\denomtnxt) + \sigma_n^2 \sigmabar_{n+1}^2 }{\sigma^2_{n+1} (\denomt) } X_0
        + \frac{\sigma^2_n    }{\denomt} X_N   \\
        =& \frac{\alpha_n^2 \sigma_{n+1}^2 + \sigmabar_{n+1}^2 (\alpha_n^2 + \sigma_n^2)  }{\sigma^2_{n+1} (\denomt) } X_0 + \frac{\sigma^2_n    }{\denomt} X_N \\
        =& \frac{\sigmabar_{n}^2}{\denomt} X_0 + \frac{\sigma^2_n    }{\denomt} X_N, \numberthis \label{eq:p6}
    \end{align*}
      where we utilize that $\sigmabar_{n}^2 + \sigma^2_n$ remains constant for all $n$ and that $\alpha_n^2 = \sigma_{n+1}^2 - \sigma_n^2 = \sigmabar_n^2 - \sigmabar_{n+1}^2$ by construction.
      Similarly, the RHS of \eqref{eq:p5} has the covariance
    \begin{align*}
        & \frac{\alpha_n^2\sigma^2_n}{\alpha_n^2 + \sigma^2_n}
        + \frac{\sigmabar_{n+1}^2\sigma_{n+1}^2}{\denomtnxt} \biggl(
             \frac{\sigma^2_n}{\alpha_n^2 + \sigma^2_n}
        \biggr)^2 \\
      =& \frac{\alpha_n^2\sigma^2_n(\denomt) + \sigmabar_{n+1}^2 \sigma_n^4 }{\sigma^2_{n+1} (\denomt) } \\
      =& \frac{\sigma_n^2 \biggl[
        \alpha_n^2 (\sigmabar_{n}^2 + \cancel{\sigma^2_{n}}) + ( \sigmabar_n^2 - \cancel{\alpha_n^2} )\sigma_n^2
       \biggr]}{\sigma^2_{n+1} (\denomt) }
       = \frac{\sigma_n^2 \sigmabar_n^2}{\sigmabar_n^2 + \sigma^2_n}. \numberthis \label{eq:p7}
    \end{align*}
    \cref{eq:p6,eq:p7} validate the equality in \eqref{eq:p5}, and we conclude the proof by induction.
\end{proof}

\begin{proof}[\textbf{Proof of \cref{prop:4}}]
    At the infinitesimal limit when $\beta_t \rightarrow 0$, the variance of $q$, \ie $\frac{\sigma_t^2 \sigmabar_t^2}{\sigmabar_t^2 + \sigma^2_t}$, vanishes as the numerator converges faster than the denominator toward zero. On the contrary, its mean remains unchanged as both ratios $(\frac{\sigmabar_t^2}{\sigmabar_t^2 + \sigma^2_t},\frac{\sigma_t^2}{\sigmabar_t^2 + \sigma^2_t})$ preserve. Hence we know the deterministic solution at the infinitesimal limit is simply $X_t = \mu_t(X_0,X_1)$.
    In this case, the diffusion of the SDE, \ie ``$\sbeta_t \rd W_t$'', vanishes while its drift approaches a vector field of the form:
    \begin{align*}
        \beta_t \gradlog \Psihat(X_t|X_0) = \frac{\beta_t}{\sigma_t^2} (X_t - X_0) := v_t(X_t| X_0).
    \end{align*}
    Hence, we have the OT-ODE in \eqref{eq:ot}.
\end{proof}

\begin{proof}[\textbf{Proof of \cref{coro:5}}]
    When $\beta_t := \beta$ is a sufficiently small constant, the ratio $\frac{\beta_t}{\sigma_t^2}$ decays in the order of $\calO(1/t)$ since $\sigma_t^2 = \int_0^t \beta_\tau \rd \tau = \beta \cdot t$. With this, \cref{prop:4} yields $\mu_t = (1-t) X_0 + t X_1$ and $v_t = \frac{X_t - X_0}{t}$. Intuitively, the OT-ODE trajectories move with a \emph{constant} velocity from  $X_1$ toward $X_0$.
\end{proof}

\section{Introduction to Schr\"odinger Bridge} \label{sec:sb-intro}

The Schr\"odinger bridge problem was originally introduced quantum mechanics \citep{schrodinger1931umkehrung,schrodinger1932theorie} and later draws broader interests with its connection to optimal transport \citep{leonard2013survey,dai1991stochastic}. The \emph{dynamic} Schr\"odinger bridge \citep{pavon1991free,leonard2012schrodinger} is typically defined as
\begin{align*}
    \min_{\mathbb{Q} \in \Pi(p_\calA,p_\calB)} \KL(\mathbb{Q} || \mathbb{P}),
\end{align*}
where $\Pi(p_\calA,p_\calB)$ is a set of path measure with the marginal densities $p_\calA$ and $p_\calB$ at the boundaries.
Relating the path measures $\mathbb{Q}$ and $\mathbb{P}$ respectively to some controlled and uncontrolled diffusion processes leads to the following stochastic optimal control (SOC) formulation:
\begin{equation}
    \begin{split} \label{eq:soc}
    &\min_{u(X_t,t)} \E\br{\int_0^1 \frac{1}{2} \norm{u(X_t,t)} \dt } \\
    \text{s.t. } \rd X_t &= [f_t(X_t) + u(X_t,t)] \dt + \sbeta \rd W_t \\
    &\qquad X_0 \sim p_\calA, \quad X_1 \sim p_\calB
\end{split}
\end{equation}
The programming \eqref{eq:soc} seeks an optimal control process $u(X_t,t)$ such that the energy cost accumulated over the time horizon $[0,1]$ is minimized while obeying the distributional boundary constraints. The coupled PDEs in \eqref{eq:sb-pde} result directly from applying the Hopf-Cole transform \citep{hopf1950partial,cole1951quasi} to the necessary conditions to \eqref{eq:soc}. This yields $u^\star(X_t,t) = \beta_t \gradlog \Psi(X_t,t)$ and hence the SDE in \eqref{eq:fsb}.
Similar reasoning applies to \eqref{eq:rsb}, where $\beta_t \gradlog \Psihat(X_t,t)$ serves as the optimal control process to a SOC similar to \eqref{eq:soc} except running backward in time.

\section{Experiment Details} \label{sec:exp-detail}

Official Pytorch implementation of our \model{} can be found in { \url{https://github.com/NVlabs/I2SB}}.

\subsection{Additional Experimental Setup}

\paragraph{Deblurring and JPEG restoration}
We adopt the implementation of blurring kernels from \citet{kawar2022denoising} and the implementation of JPEG quality factor from \citet{kawar2022jpeg}.
Following the baselines \citep{saharia2022palette,anonymous}, the FID is evaluated over the reconstruction results on the 10k ImageNet validation subset,\footnote{
    \url{https://bit.ly/eval-pix2pix} \label{footnote:palette}
} and compared against the statistics of the entire ImageNet validation set.

\paragraph{4$\times$ super-resolution}
We adopt the same implementation of filters from DDRM \citep{kawar2022denoising}.
We first generate 64$\times$64 images then upsample them to 256$\times$256 before passing into \model{}, since the model transports between clean and degraded images of the same size.
Following the baselines \citep{saharia2022palette,anonymous}, the FID is evaluated over the reconstruction results on the entire ImageNet validation set, and compared against the statistics of the entire ImageNet training set.

\paragraph{Inpainting}

We use the same freeform masks provided by Palette \citep{saharia2022palette},$^\text{\ref{footnote:palette}}$ which contains 10000 masks for both \textit{10\%-20\%} and \textit{20\%-30\%} ratios.
We randomly select these masks during training and iterate them through the 10k ImageNet validation subst$^\text{\ref{footnote:palette}}$ for reproducible evaluation.
We follow the same instructions from Palette and set up \model{} such that
\textit{(i)} the training loss is restricted to only the masked regions,
\textit{(ii)} the masked regions are filled with Gaussian noise as inputs (see \cref{fig:app-inpaint}), and
\textit{(iii)} the model predicts only the masked regions during generation.
The FID is evaluated over the reconstruction results on the 10k ImageNet validation subset and compared against the statistics of the entire ImageNet validation set.

\paragraph{Evaluation}
We use \texttt{cleanfid} package\footnote{
    \url{https://github.com/GaParmar/clean-fid}
} with the option ``\texttt{legacy\_pytorch}'' to compute FID values.
For the reference statistics, we take the ones provided by ADM \citep{dhariwal2021diffusion} for the ImageNet training set and compute the ones for the ImageNet validation set by resizing and center-cropping the images to 256$\times$256, similar to ADM.
The Classifier Accuracy is based on a pre-trained ResNet50 \citep{he2016deep}.
Following the suggestions from \citet{saharia2022palette}, we avoid pixel-level metrics like PSNR and SSIM as they tend to prefer blurry regression outputs \citep{menon2020pulse,ledig2017photo}.

\paragraph{Palette implementation}

We implement our own Palette for the results in \cref{table:deblur,table:inpaint,table:gan,fig:nfe-curve,fig:nfe-visual}. For all the other tasks, we use the official values reported in their paper.
For a fair comparison,
we initialize its network of Palette with the same checkpoint from unconditional ADM~\citep{dhariwal2021diffusion} on ImageNet 256$\times$256 and concatenate the first layer with conditional inputs, following \citet{rombach2022high}. The SDE uses the same 1000 time steps with quadratic discretization similar to \model{}.

\subsection{Additional Qualitative Results}

\begin{figure*}[t]
    \vskip 0.05in
    \begin{center}
    \centerline{\includegraphics[width=\textwidth]{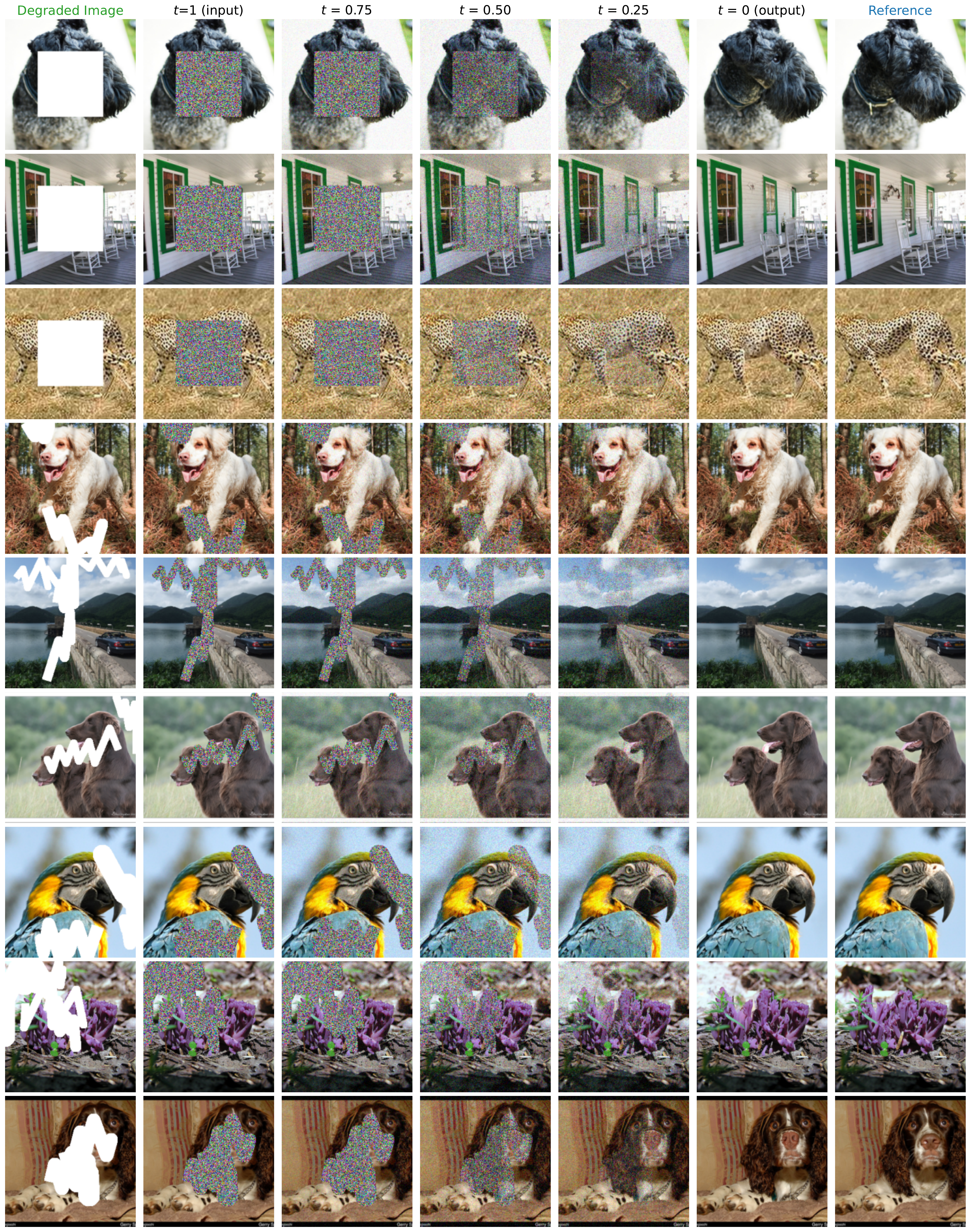}}
    \vskip -0.1in
    \caption{
        Generative processes of \model{} on inpainting tasks.
        \textbf{Top 3 rows}: \textit{Center 128$\times$128} mask.
        \textbf{Middle 3 rows}: \textit{Freeform 10\%-20\%} mask.
        \textbf{Bottom 3 rows}: \textit{Freeform 20\%-30\%} mask.
    }
    \label{fig:app-inpaint}
    \end{center}
    \vskip -0.3in
\end{figure*}

\begin{figure*}[t]
    \vskip 0.05in
    \begin{center}
    \centerline{\includegraphics[width=\textwidth]{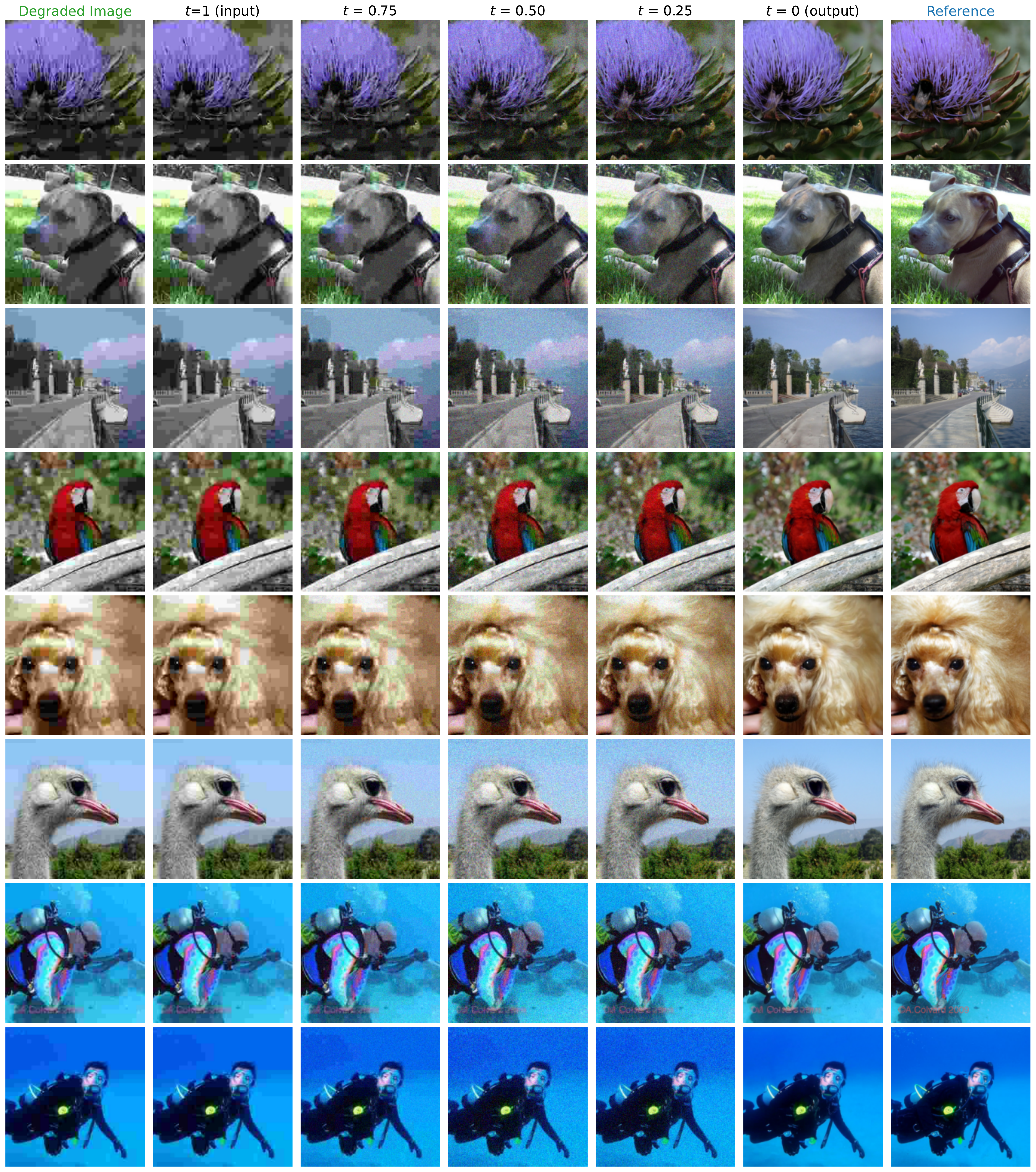}}
    \vskip -0.1in
    \caption{
        Generative processes of \model{} on JPEG restoration tasks.
        \textbf{Top 5 rows}: QF=5.
        \textbf{Bottom 3 rows}: QF=10.
    }
    \label{fig:app-jpeg}
    \end{center}
    \vskip -0.3in
\end{figure*}

\begin{figure*}[t]
    \vskip 0.05in
    \begin{center}
    \centerline{\includegraphics[width=\textwidth]{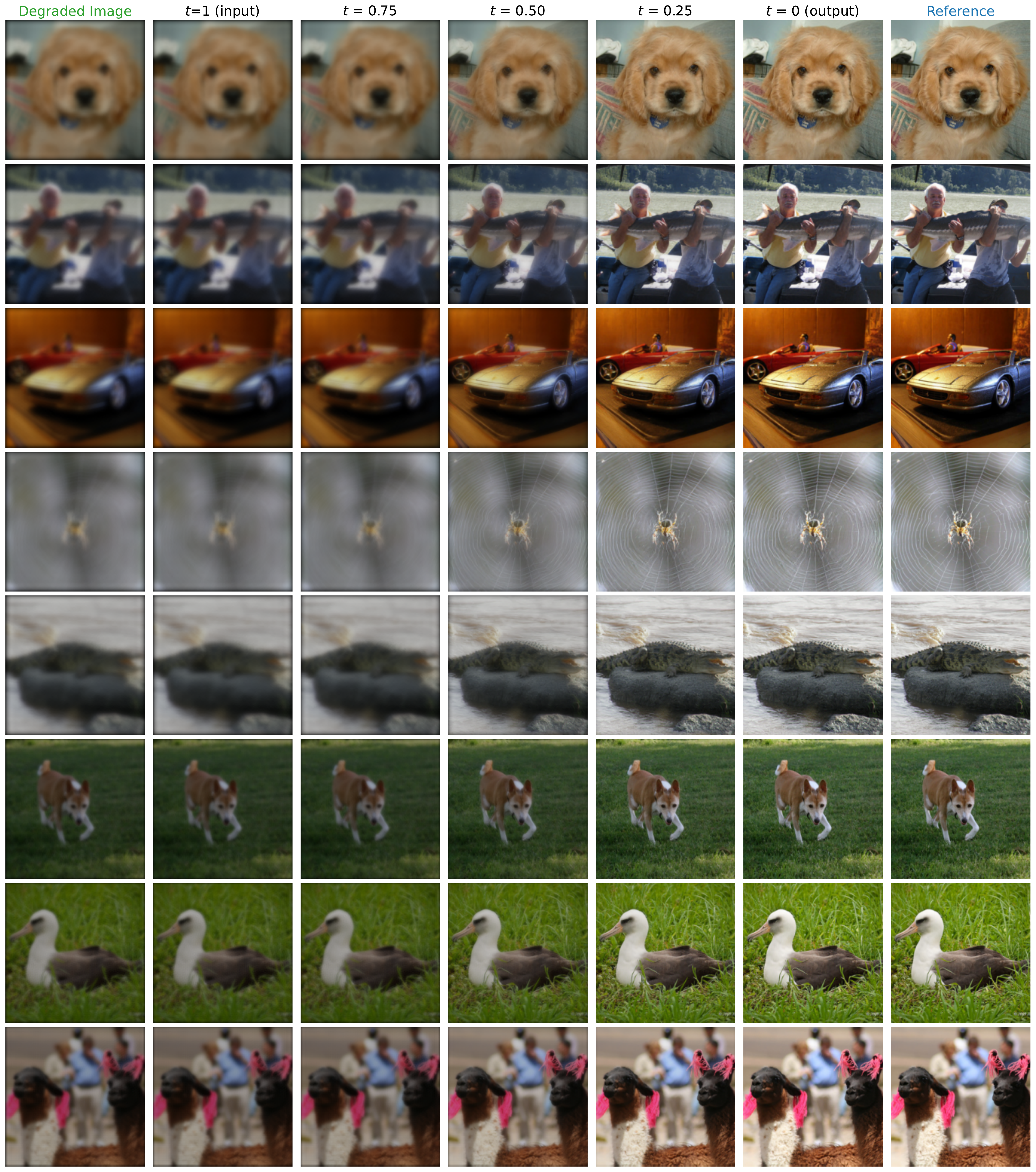}}
    \vskip -0.1in
    \caption{
        Generative processes of \model{} on deblurring tasks.
        \textbf{Top 5 rows}: \textit{Uniform} kernel.
        \textbf{Bottom 3 rows}: \textit{Gaussian} kernel.
    }
    \label{fig:app-deblur}
    \end{center}
    \vskip -0.3in
\end{figure*}

\begin{figure*}[t]
    \vskip 0.05in
    \begin{center}
    \centerline{\includegraphics[width=\textwidth]{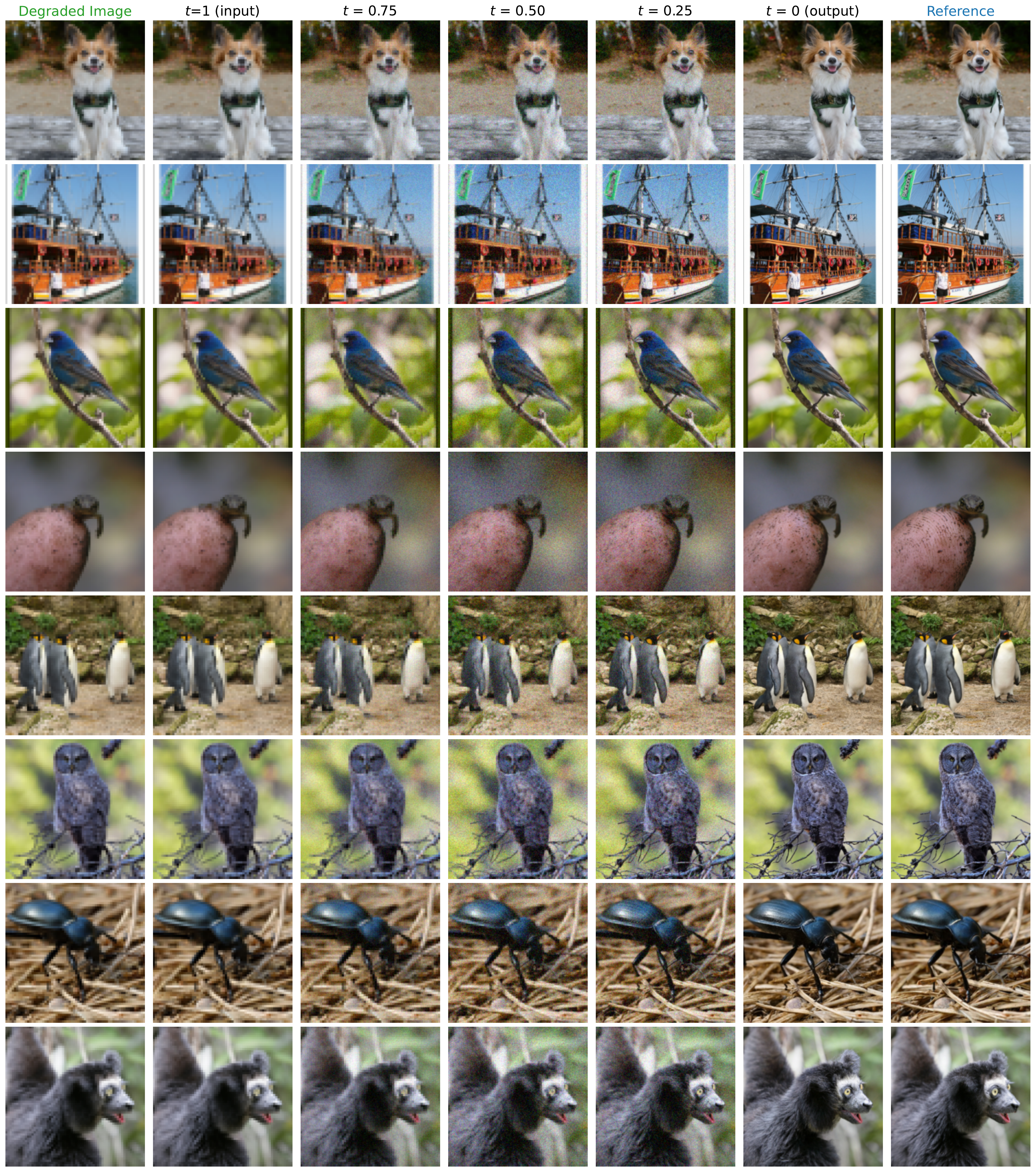}}
    \vskip -0.1in
    \caption{
        Generative processes of \model{} on 4$\times$ super-resolution tasks.
        \textbf{Top 5 rows}: \textit{Pool} filter.
        \textbf{Bottom 3 rows}: \textit{Bicubic} filter.
    }
    \label{fig:app-sr4x}
    \end{center}
    \vskip -0.3in
\end{figure*}

\begin{figure*}[t]
    \vskip 0.05in
    \begin{center}
    \centerline{\includegraphics[width=\textwidth]{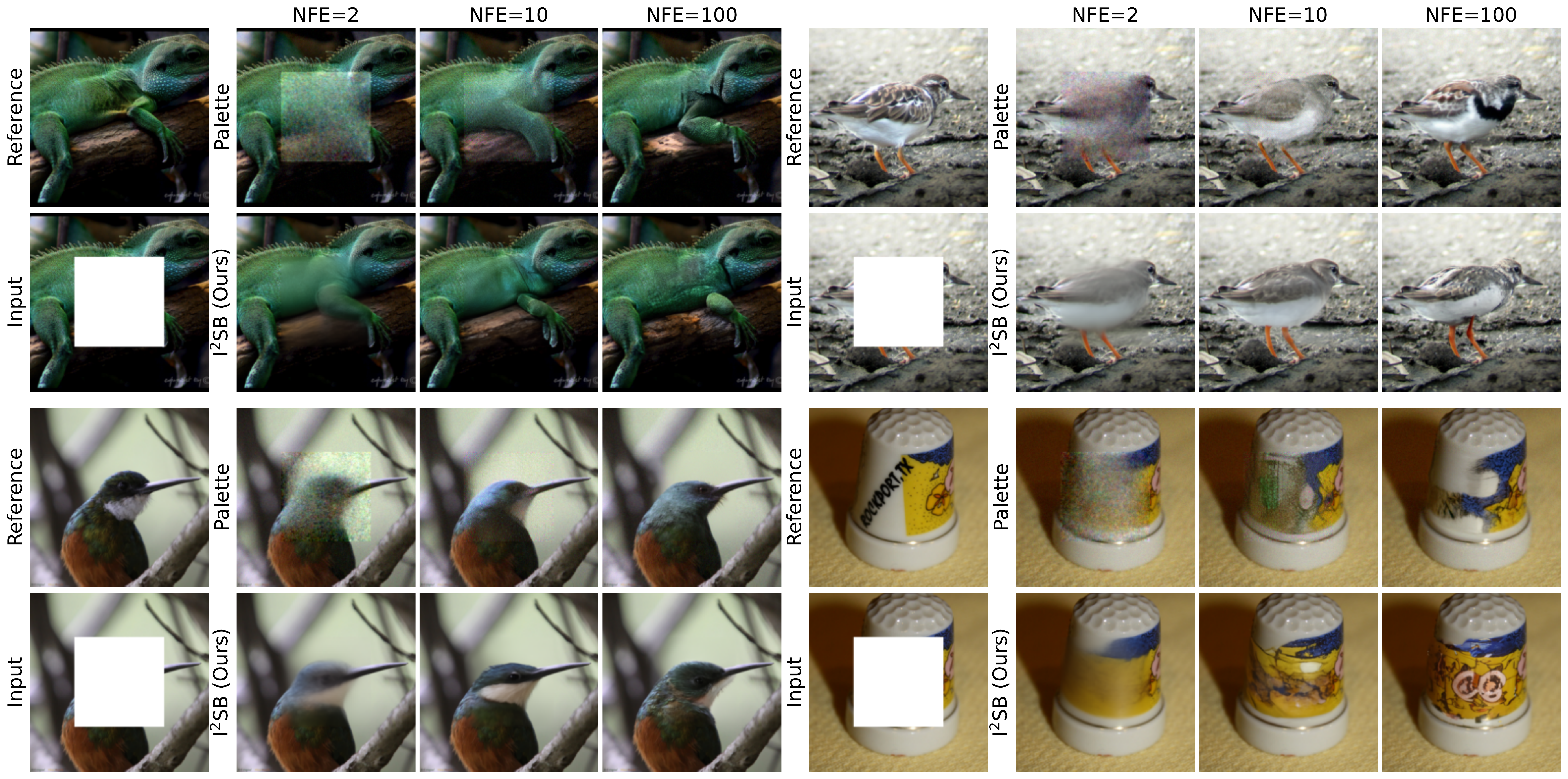}}
    \vskip -0.1in
    \caption{
        Additional qualitative comparison between \model{} and Palette on inpainting (\textit{Center 128$\times$128}).
    }
    \label{fig:app-nfe-inpc}
    \end{center}
    \vskip -0.3in
\end{figure*}

\begin{figure*}[t]
    \vskip 0.05in
    \begin{center}
    \centerline{\includegraphics[width=\textwidth]{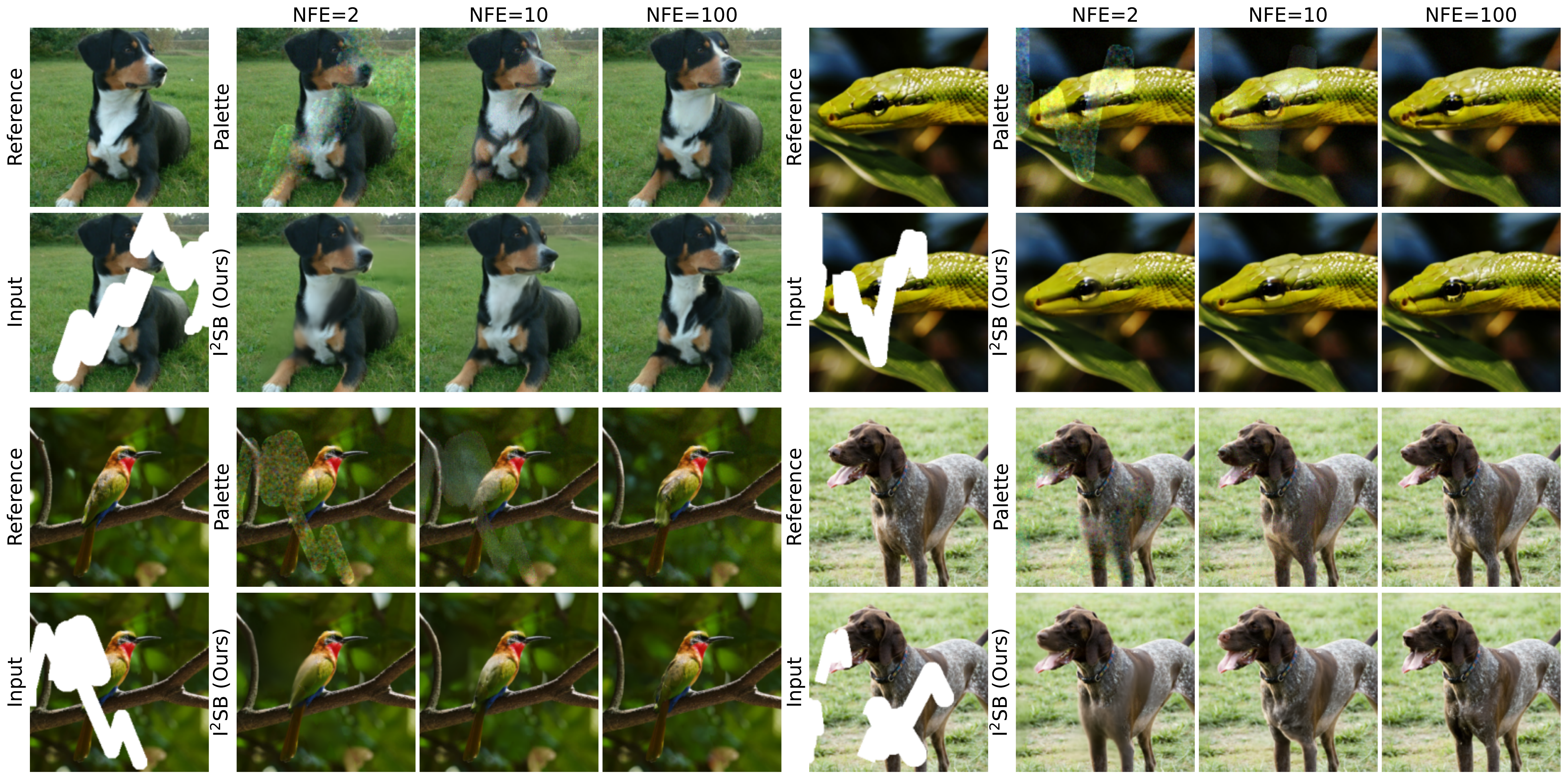}}
    \vskip -0.1in
    \caption{
        Additional qualitative comparison between \model{} and Palette on inpainting (\textit{Freeform 20\%-30\%}).
    }
    \label{fig:app-nfe-inpf}
    \end{center}
    \vskip -0.3in
\end{figure*}

\begin{figure*}[t]
    \vskip 0.05in
    \begin{center}
    \centerline{\includegraphics[width=\textwidth]{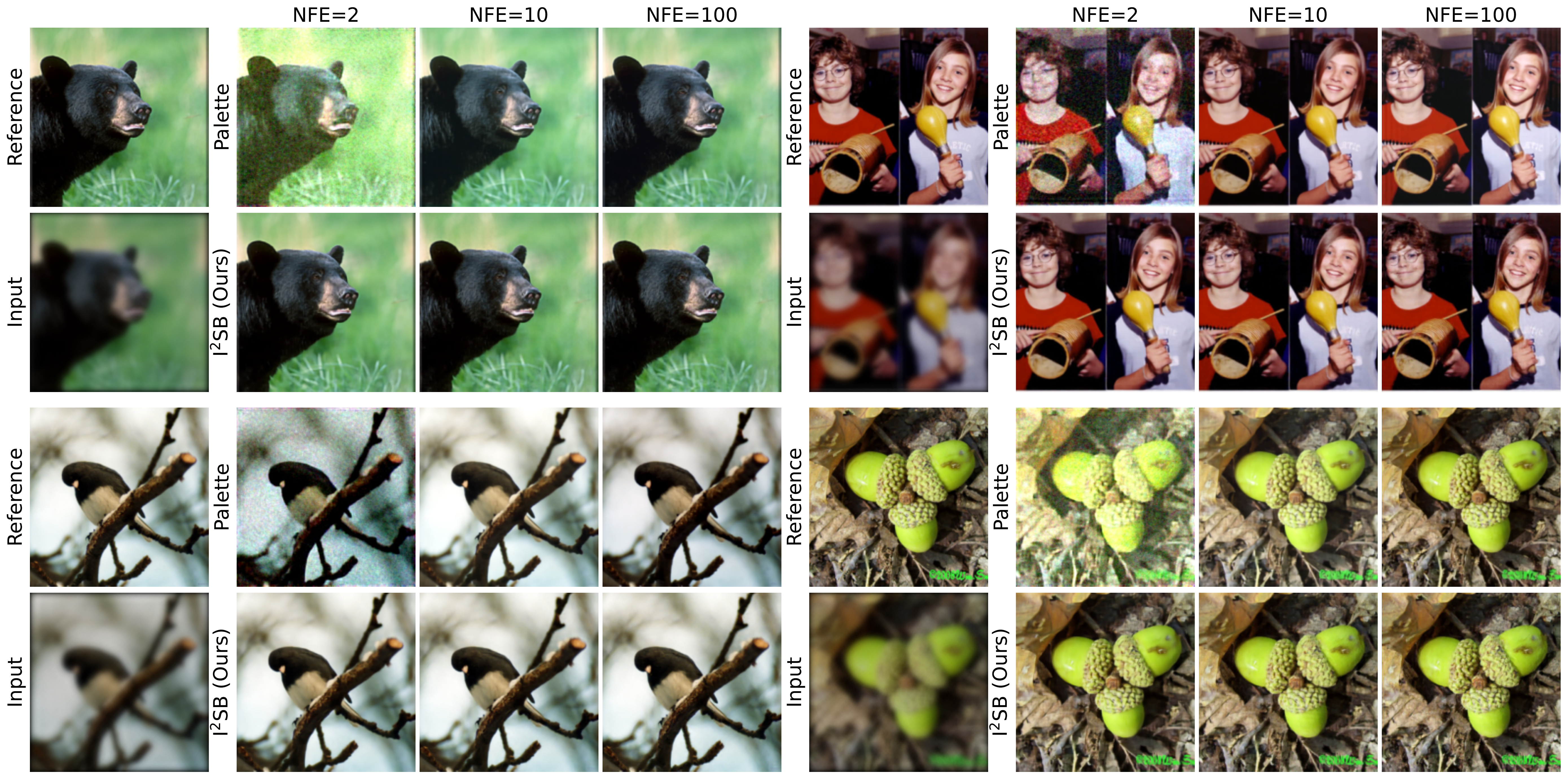}}
    \vskip -0.1in
    \caption{
        Additional qualitative comparison between \model{} and Palette on deblurring (\textit{Uniform}).
    }
    \label{fig:app-nfe-bluru}
    \end{center}
    \vskip -0.3in
\end{figure*}

\begin{figure*}[t]
    \vskip 0.05in
    \begin{center}
    \centerline{\includegraphics[width=\textwidth]{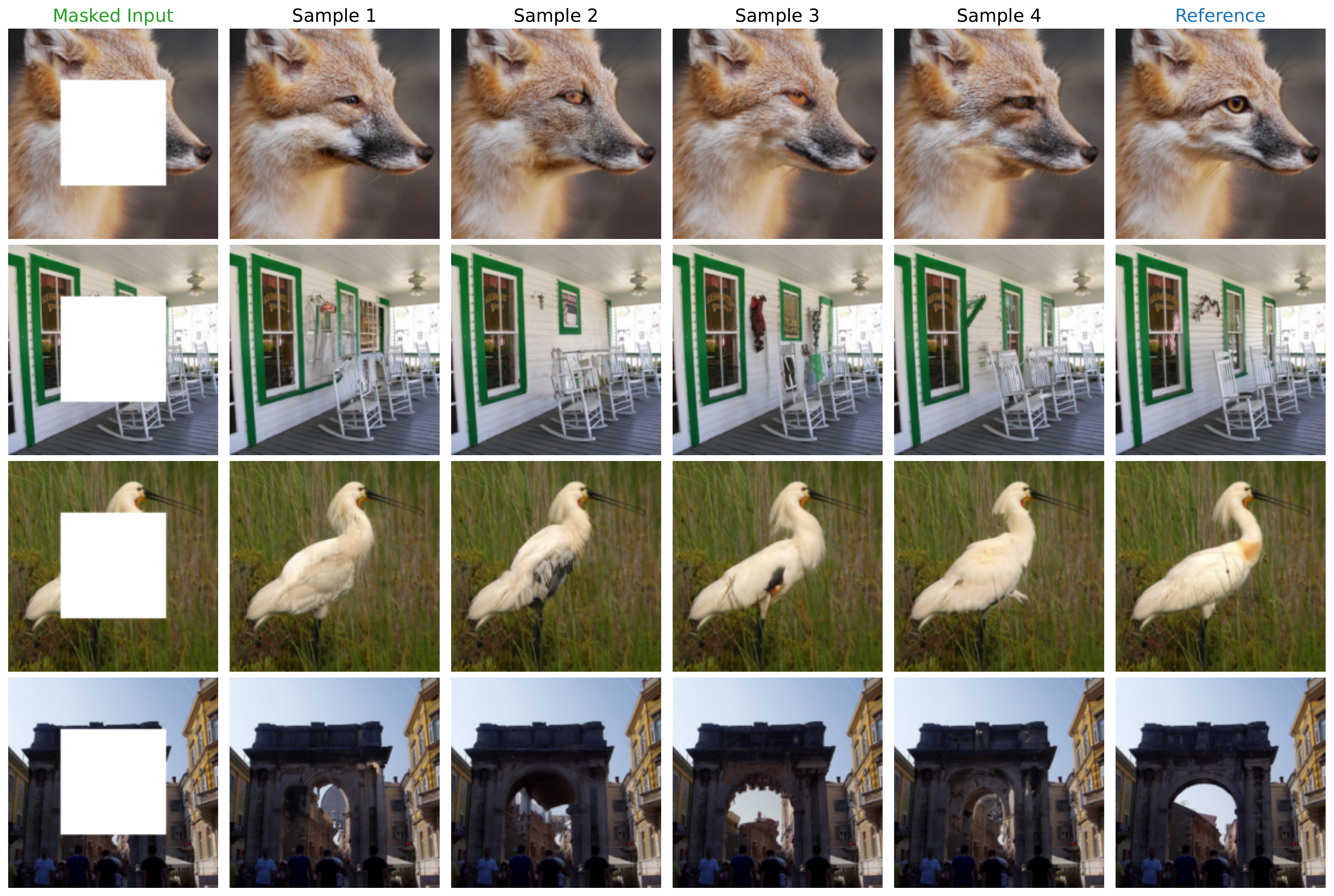}}
    \vskip -0.1in
    \caption{
        Diversity of \model{} outputs on inpainting tasks.
    }
    \label{fig:app-inp-diversity}
    \end{center}
    \vskip -0.3in
\end{figure*}

\cref{fig:app-inpaint,fig:app-jpeg,fig:app-sr4x,fig:app-deblur} provide additional qualitative results on each restoration tasks, and \cref{fig:app-nfe-bluru,fig:app-nfe-inpc,fig:app-nfe-inpf} provide additional examples comparing between Palette and \model{} w.r.t. various NFE sampling. Finally, \cref{fig:app-inp-diversity} demonstrates that \model{} is able to generate diverse samples.

\subsection{Additional Discussions} \label{sec:more-discuss}

\begin{table}[t]
    \caption{
        Additional ablation study on the effect of stochasticity on inpainting tasks. OT-ODE exhibits severe degradation with \emph{noiseless} masks but yields slightly better results after injecting additional noise to the masked regions of degraded inputs.
    }
    \label{table:aba-ot2}
    \begin{center}
    \begin{small}
    \begin{tabular}{ccccc}
    \toprule
    & \multicolumn{2}{c}{mask} & \multicolumn{2}{c}{mask + noise} \\[0.5pt]
    & \textit{Center} & \textit{Ff. 20-30\%} & \textit{Center} & \textit{Ff. 20-30\%} \\
    \midrule
    FID diff. & \markred{+50.9} & \markred{+13.0} & \markgreen{-0.1} & \markgreen{-0.1} \\[0.5pt]
    CA  diff. & \markred{-14.1} & \markred{-7.3} & {0.0} & \markgreen{+0.7} \\
    \bottomrule
    \end{tabular}
    \end{small}
    \end{center}
    \vskip -0.05in
\end{table}

\paragraph{More Ablation Study on OT-ODE}

\cref{table:aba-ot} shows how OT-ODE seems to disfavor restoration tasks with large uncertainties.
We conjecture that it is due to the severe information lost in degraded inputs that hinders the reconstruction of deterministic mapping.
This is validated in \cref{table:aba-ot2}, where we compare the performance difference on inpainting tasks with or without injecting Gaussian noise to the masked regions.
OT-ODE exhibits severe degradation without any stochasticity but yields comparable results after injecting additional noise to the masked regions of degraded inputs.

\paragraph{Other Parameterization} \label{sec:edm-param}

\begin{figure}[t]
    \vskip 0.05in
    \begin{center}
    \centerline{\includegraphics[width=\columnwidth]{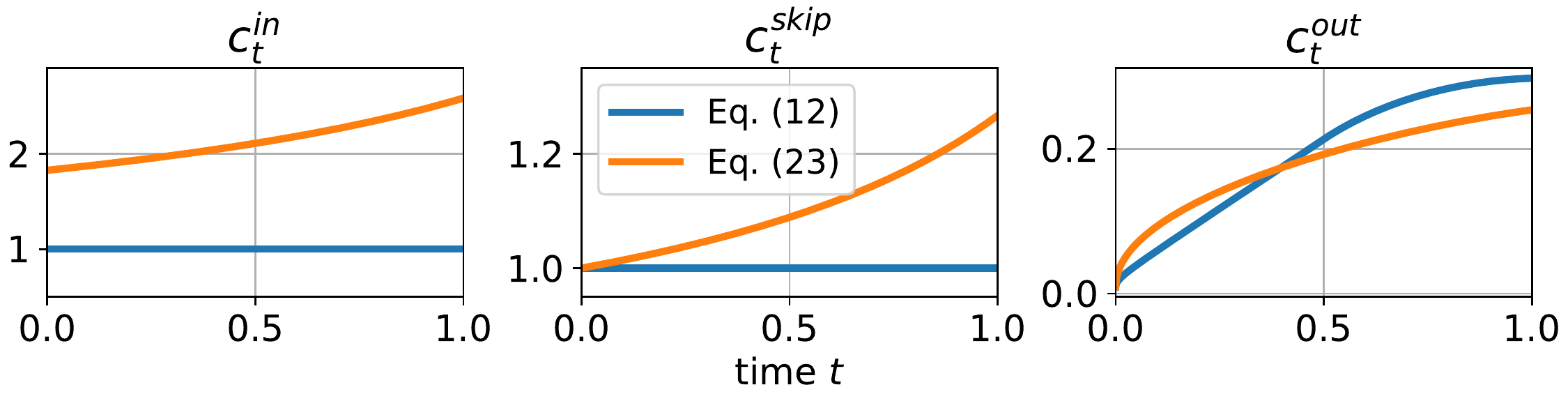}}
    \vskip -0.15in
    \caption{
        The numerical values of the coefficients $\cin$, $\cskip$, $\cout$ adopted in (\ref{eq:obj}, \ref{eq:coef-edm}) for training $\epsilon(X_t,t;\theta)$, where $X_t$ interpolates between clean and corrupted image pair $(X_0,X_1)$.
    }
    \label{fig:4}
    \end{center}
    \vskip -0.05in
\end{figure}

\def\ccin{{ c_\text{in} }}
\def\ccout{{ c_\text{out} }}
\def\ccskip{{ c_\text{skip} }}

In additional to the standard rescaled score function in \eqref{eq:obj}, we may follow \citet{karras2022elucidating} by considering
\begin{align} \label{eq:obj2}
    \norm{
        \epsilon(\cin \cdot X_t,t; \theta) - \frac{\cskip \cdot X_t - X_0}{\cout}
    },
\end{align}
where $\cin, \cskip, \cout \in \sR$ are time-varying coefficients such that \textit{(i)} the inputs and outputs of $\epsilon$ have unit variance and \textit{(ii)} the approximation error induced from $\epsilon$ are minimized.
In our cases, since $X_t$ now interpolates between clean and corrupted image pairs (rather than images with i.i.d. noises),
we re-derive these coefficients in a more general form given estimated $\Var[X_t]$ and $\Cov[X_0, X_t]$:
\begin{equation} \label{eq:coef-edm}
    \begin{split}
    \cin = \tfrac{1}{\sqrt{\Var[X_t]}}, \quad
    \cskip = \tfrac{\Cov[X_0, X_t]}{\Var[X_t]}, \\
    \cout = \sqrt{ \Var[X_0] - \tfrac{\Cov[X_0, X_t]^2}{\Var[X_t]} }.
    \end{split}
\end{equation}
These coefficients can be obtained by $\Var[\ccin X_t] = 1$ and
\begin{align*}
    &~\Var\br{\frac{\ccskip X_t - X_0}{\ccout}} = 1 \\
    \Rightarrow&~c_\text{skip}^2 \Var[X_t] + \Var[X_0] - 2 \ccskip \Cov[X_t,X_0] = c_\text{out}^2.
\end{align*}
Choosing $\ccskip$ such that $c_\text{out}^2$ is minimized yields \eqref{eq:coef-edm}.
\cref{fig:4} summarizes the difference between \eqref{eq:obj} and \eqref{eq:coef-edm}.
In practice, we find their empirical differences negligible.

\begin{remark}[How \eqref{eq:coef-edm} recovers \citet{karras2022elucidating}]
    In the specific case when $X_t := X_0 + \epsilon$,
    $X_0$ has variance $\sigma_{\text{data}}^2$, and $\epsilon$ is i.i.d. noise with variance $\sigma^2$,
    we have
    \begin{equation}
    \begin{split} \label{eq:edm2}
        \Var[X_t] &= \sigma_{\text{data}}^2 + \sigma^2 \\
        \Cov[X_0,X_t] &= \Var[X_0] + \Cov[X_0,\epsilon] = \sigma_{\text{data}}^2.
    \end{split}
    \end{equation}
    Substituting \eqref{eq:edm2} into \eqref{eq:coef-edm} yields the coefficients suggested in \citet{karras2022elucidating}.
\end{remark}

\end{document}